\documentclass[11pt]{article}

\usepackage[preprint]{acl}

\usepackage{times}
\usepackage{latexsym}
\usepackage{amsmath}
\usepackage{amssymb}
\usepackage{booktabs}
\usepackage{makecell}
\usepackage{multirow}
\usepackage{colortbl}

\definecolor{goodproperty}{RGB}{224, 242, 225}
\definecolor{badproperty}{RGB}{252, 226, 226}
\definecolor{partialproperty}{RGB}{253, 243, 208}
\definecolor{notlistedprop}{RGB}{240, 241, 243}

\newcommand{\goodprop}[1]{\cellcolor{goodproperty}#1}
\newcommand{\badprop}[1]{\cellcolor{badproperty}#1}
\newcommand{\partialprop}[1]{\cellcolor{partialproperty}#1}
\newcommand{\greyprop}[1]{\cellcolor{notlistedprop}\textcolor{gray}{#1}}
\newcommand{\scite}[1]{{\footnotesize\citep{#1}}}

\newcommand\nnfootnote[1]{%
  \begin{NoHyper}
  \renewcommand\thefootnote{}\footnote{#1}%
  \addtocounter{footnote}{-1}%
  \end{NoHyper}
}

\usepackage[T1]{fontenc}

\usepackage[utf8]{inputenc}

\usepackage{microtype}

\usepackage{inconsolata}

\usepackage{graphicx}

\definecolor{poyColor}{RGB}{1, 115, 113}  
\usepackage{todonotes}

\newtheorem{proposition}{Proposition}

\newcommand{\gemmu}{\textbf{GEMMU} }

\title{Are We Really Measuring Forgetting? Pitfalls in Evaluating Machine Unlearning for Multimodal LLMs}

\author{Oscar Chew$^{*,1}\quad$Po-Yi Lu$^{*,1,2}\quad$Yi-Jie Cheng$^3\quad$Kunal Jain$^1$\\
\bf Hsuan-Tien Lin$^2\quad$Zhengzhong Tu$^1\quad$Kuan-Hao Huang$^1$ \vspace{0.1em}\\
  $^1$Texas A\&M University$\quad^2$National Taiwan University$\quad^3$Columbia University \vspace{0.1em}\\
  \texttt{\{oscarchew,kjain,tzz,khhuang\}@tamu.edu} \\
  \texttt{\{d09944015,htlin\}@csie.ntu.edu.tw \ \ yc4708@columbia.edu}}

\begin{document}
\maketitle
\begin{abstract}
As machine unlearning for multimodal large language models (MLLMs) rapidly advances, rigorous evaluation of its effectiveness becomes increasingly important. However, existing benchmarks largely inherit evaluation protocols from text-only unlearning and overlook the unique challenges of unlearning in multimodal settings. In this work, we identify three major pitfalls that may lead to misleading conclusions: the lack of heldout evaluation to distinguish genuine concept-wise forgetting from instance-wise overfitting, the reliance on synthetic, fine-tuned knowledge rather than real-world knowledge acquired during pre-training, and unfair method comparisons caused by applying the same training settings to different unlearning methods. To address these challenges, we introduce \textbf{GEMMU} (\textbf{G}eneralizable \textbf{E}valuation for \textbf{M}ultimodal \textbf{M}achine \textbf{U}nlearning), a benchmark built on real-world knowledge and a principled evaluation protocol. Re-evaluating representative unlearning methods under \gemmu reveals substantial generalization gaps and challenges conclusions drawn from previous benchmarks. Our findings highlight the importance of rigorous evaluation for advancing multimodal machine unlearning and provide guidelines for future benchmark design.
\end{abstract}
\section{Introduction}
\nnfootnote{$^{*}$ denotes equal contributions.}
As MLLMs \citep{bai2025qwen3,li2025llavanextinterleave} are increasingly deployed in real-world applications, concerns surrounding privacy and data governance have intensified. In this context, machine unlearning has attracted growing attention as a mechanism for removing sensitive or private knowledge, such as personal information, from trained models. As unlearning methods advance, rigorous evaluation is essential to reliably assess their effectiveness.

Several benchmarks \citep{liu2025protecting,dontsov2025clear,wang2025umu} have been proposed to assess progress in multimodal machine unlearning. However, we identify the following fundamental pitfalls in existing benchmarks that may lead to overly optimistic or misleading evaluations.

\underline{First}, testing on the training set is widely recognized as poor practice in machine learning, yet existing unlearning protocols often evaluate forgetting and retention on the same data used during unlearning \cite{liu2025protecting,ma2025benchmarking,zheng2026offside}. Such evaluation risks overestimating unlearning effectiveness, as performance on these examples may not generalize to unseen data. This concern is particularly important in the multimodal setting, where effective unlearning should remove knowledge at the \textit{concept level} rather than merely suppressing specific visual \textit{instances}.

\underline{Second}, many existing benchmarks \citep{dontsov2025clear,wang2025umu,liu2025protecting,xu2025pebench} focus primarily on newly fine-tuned knowledge, rather than real-world knowledge acquired during pre-training. While convenient for evaluation, such settings may not reflect realistic unlearning scenarios. Moreover, existing unlearning methods largely rely on reverse fine-tuning, which may be more effective at removing newly introduced knowledge. In contrast, pre-trained knowledge is often deeply entangled with other useful knowledge, making it difficult to remove selectively without affecting the model's general capabilities. As a result, those benchmarks may overestimate the effectiveness of existing methods in real-world unlearning.

\begin{figure*}
    \centering
    \includegraphics[width=\linewidth]{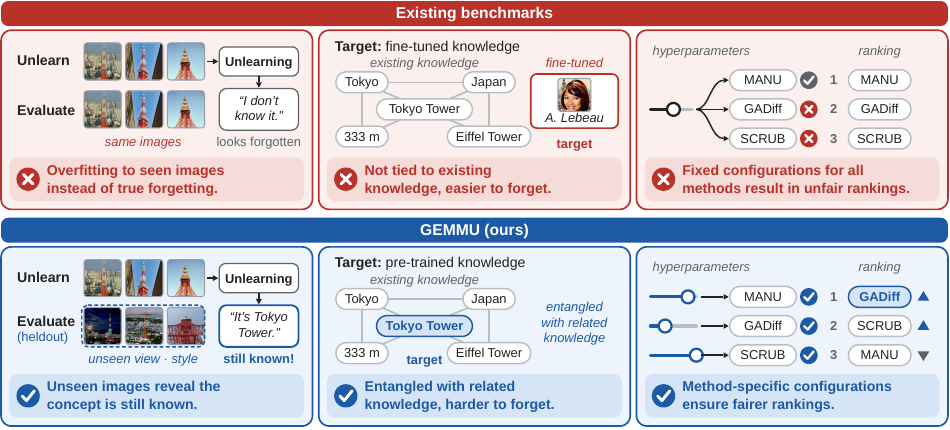}
    \caption{Three major pitfalls in existing multimodal unlearning benchmarks (top) and how GEMMU addresses them (bottom) through heldout evaluation, real-world knowledge unlearning, and method-specific hyperparameter tuning to achieve generalizable evaluations.}
    \label{fig:gemmu}
\end{figure*}

\underline{Third}, most existing benchmarks evaluate different unlearning methods using identical hyperparameters, despite differences in their optimization objectives and training dynamics \cite{xu2025pebench,dontsov2025clear}. A configuration that works well for one method may be suboptimal for another. Consequently, conclusions about which method is more effective can depend on the shared configuration rather than on the methods themselves.

To address these limitations, we introduce \textbf{GEMMU} (Generalizable Evaluation for Multimodal Machine Unlearning; Fig.~\ref{fig:gemmu}), a benchmark built on three principles: heldout evaluation for both forget and retain sets, unlearning of real-world pre-trained knowledge, per-method hyperparameter search (Tab. ~\ref{tab:benchmarks_comparisons}), and additionally modality gap analysis (Sec.~\ref{sec:modality_gap}). Re-evaluating representative unlearning methods with \gemmu reveals substantial generalization gaps overlooked by existing benchmarks, shows that method rankings can change under different configurations, and calls into question the effectiveness of current methods. Our main contributions are summarized as follows:
\begin{itemize}
    \item We identify major pitfalls in existing MLLM unlearning evaluations: \textit{testing on the training set}, \textit{reliance on fine-tuned knowledge}, and \textit{unfair hyperparameter settings} across methods.
    \item We introduce \textbf{GEMMU}, a more rigorous benchmark that addresses these limitations and provides principles for future benchmark design.
    \item We re-evaluate nine representative unlearning methods and uncover substantial generalization gaps and ranking changes overlooked by previous benchmarks.
\end{itemize}

\section{Related Works}
\paragraph{Machine Unlearning Methods for MLLMs}
\label{sec:unlearning_methods}
Existing approaches to multimodal machine unlearning mainly follow a \emph{reverse fine-tuning} paradigm, typically suppressing forget-set knowledge through gradient ascent while preserving retained knowledge through gradient descent or other regularization objectives. Representative methods include GA \citep{thudi2022unrolling}, GADiff \citep{liu2022continual}, KL Min and IDK \citep{maini2024tofu}, NPO \citep{zhang2024negative}, LLMU \citep{yao2024large}, RMU \citep{li2024wmdp}, and SCRUB \citep{kurmanji2023towards}. While some of the above methods were originally developed for text-only LLMs, they have served as common baselines for unlearning in MLLMs. More recently, model editing has emerged as an alternative to conventional tuning-based unlearning. For example, MANU \citep{liu2025modality} selectively identifies and prunes model components associated with the knowledge to be forgotten.

\begin{table*}[]
    \centering
    \resizebox{\linewidth}{!}{%
    \begin{tabular}{lccccccc}
    \toprule
    Benchmark & \makecell{Proper \\ Train-test Split} \footnotemark& \makecell{Knowledge \\ Acquisition} & \makecell{Unlearned \\ Target Source} & \makecell{\# Real-world \\ Concepts} & \makecell{Concept \\ Types} & \makecell{Hyperparameter \\ Selection} & Open-source \\
    \toprule
    MMUBench \scite{li2024single} & \partialprop{Partial} & \goodprop{Pre-trained} & \goodprop{Real-world} & \goodprop{20} & \goodprop{Diverse} & \badprop{One-size-fits-all} & \badprop{No}\\
    MLLMU-Bench \scite{liu2025protecting} & \partialprop{Partial} & \badprop{Fine-tuned} & \badprop{Synthetic} & \badprop{--} & \badprop{Human only} & \badprop{One-size-fits-all} & \goodprop{Yes}\\
    FIUBench \scite{ma2025benchmarking} & \badprop{No} & \badprop{Fine-tuned} & \badprop{Synthetic} & \badprop{--} & \badprop{Human only} & \partialprop{Per-method} & \goodprop{Yes}\\
    CLEAR \scite{dontsov2025clear} & \badprop{No} & \badprop{Fine-tuned} & \badprop{Synthetic} & \badprop{--} & \badprop{Human only} & \badprop{One-size-fits-all} & \goodprop{Yes}\\
    PEBench \scite{xu2025pebench} & \partialprop{Partial} & \badprop{Fine-tuned} & \badprop{Synthetic} & \badprop{--} & \partialprop{Humans, events} & \partialprop{Per-method} & \goodprop{Yes} \\
    UMU-Bench \scite{wang2025umu} & \badprop{No} & \badprop{Fine-tuned} & \badprop{Synthetic} & \badprop{--} & \badprop{Human only} & \badprop{One-size-fits-all} & \goodprop{Yes}\\
    PULSE \scite{kawakami2025pulse} & \badprop{No} & \goodprop{Pre-trained} & \goodprop{Both} & \goodprop{20} & \badprop{Human only} & \greyprop{Not listed} & \badprop{No} \\
    OFFSIDE \scite{zheng2026offside} & \badprop{No} & \badprop{Fine-tuned} & \goodprop{Real-world} & \goodprop{80} & \badprop{Human only} & \badprop{One-size-fits-all} & \goodprop{Yes}\\
    \midrule
    \gemmu (Ours) & \goodprop{Yes} & \goodprop{Pre-trained} & \goodprop{Both} & \goodprop{\textbf{139}} & \goodprop{Diverse} & \goodprop{Searched} & \goodprop{Yes}\\
    \bottomrule
    \end{tabular}%
    }
    \caption{Comparison of existing multimodal machine unlearning benchmarks. \gemmu is the most comprehensive benchmark that evaluates the unlearning of real-world, pre-trained knowledge with a proper train-test split.}
    \label{tab:benchmarks_comparisons}
\end{table*}
\begin{figure}
    \centering
    \includegraphics[width=0.9\linewidth]{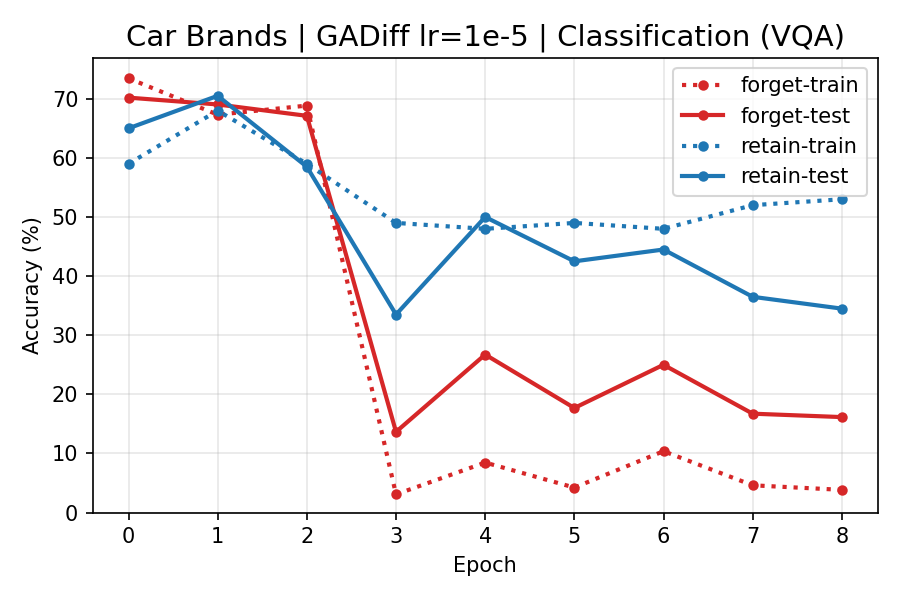}
    \caption{A representative example of the generalization gap. The red dotted curve falls well below the red solid curve, while the blue dotted curve remains above the blue solid curve, revealing substantial train–test discrepancies after unlearning.}
    \label{fig:generalization gap}
    \vspace{-1em}
\end{figure}

\paragraph{Machine Unlearning Evaluation for MLLMs}
Early language-only benchmarks such as TOFU~\citep{maini2024tofu} establish a controlled evaluation setting for fictitious knowledge removal, and MLLMU-Bench~\citep{liu2025protecting} extends this direction to multimodal models. CLEAR~\citep{dontsov2025clear}, FIUBench~\citep{ma2025benchmarking}, UMU-Bench~\citep{wang2025umu}, and PEBench~\citep{xu2025pebench} introduce more comprehensive multimodal evaluation settings covering person profiles, cross-modal knowledge, and personal entities with event scenes. Despite this progress, existing benchmarks remain limited for evaluating practical multimodal unlearning. They still rely on generated identities, which require fine-tuning to memorize this new knowledge and may not reflect the unlearning of real-world, pre-trained knowledge. MMUBench and PEBench \citep{li2024single,xu2025pebench} partition the forget set into training and test examples but overlook the same problem for the retain set. An existing preprint, PULSE \citep{kawakami2025pulse}, studies the unlearning of pre-trained knowledge, but does not provide a theoretical analysis of the distinction between fine-tuned and pre-trained knowledge, nor does it address overfitting to the forget or retain sets.
\nnfootnote{$^1$The entry is No if the benchmark reports forget/retain metrics evaluated on the same data split used for optimization.}
\section{Preliminaries}
\paragraph{Formulation of the Unlearning Task}
Let $f_\theta$ denote an MLLM with parameters $\theta$. Following the standard formulation in this field, the unlearning task considers two datasets: a forget set $\mathcal{D}_{\text{forget}}$, containing knowledge to be removed, and a retain set $\mathcal{D}_{\text{retain}}$, containing knowledge to be preserved. Each dataset consists of image-question-answer triplets $(I, q, a)$, where $I$ is empty for text-only questions. We further partition both datasets into disjoint training and test sets to evaluate generalizability: $\mathcal{D}_{\text{forget}}^{\text{train}}, \mathcal{D}_{\text{forget}}^{\text{test}}, 
\mathcal{D}_{\text{retain}}^{\text{train}}
$ and $\mathcal{D}_{\text{retain}}^{\text{test}}$.

Given a pre-trained MLLM $f_{\theta_0}$, where $\theta_0$ denotes the parameters before unlearning, an unlearning method $\mathcal{U}$ updates $\theta_0$ using the training splits:
\begin{equation*}
    \theta_u = \mathcal{U}(
        \theta_0,
        \mathcal{D}_{\text{forget}}^{\text{train}},
        \mathcal{D}_{\text{retain}}^{\text{train}}
    ),
\end{equation*}
where $\theta_u$ denotes the parameters after unlearning. Note that previous works often use the entire $\mathcal{D}_{\text{forget}}$ and $\mathcal{D}_{\text{retain}}$ in place of the training splits mentioned above. The objective is to reduce the model's ability to recover knowledge in $\mathcal{D}_{\text{forget}}$, while preserving its performance on $\mathcal{D}_{\text{retain}}$. Successful unlearning should achieve these goals not only on the training splits but also on the test splits, which provide a more reliable measure of generalizable forgetting and knowledge retention.

\begin{figure*}[]
    \centering
    \includegraphics[width=0.47\linewidth]{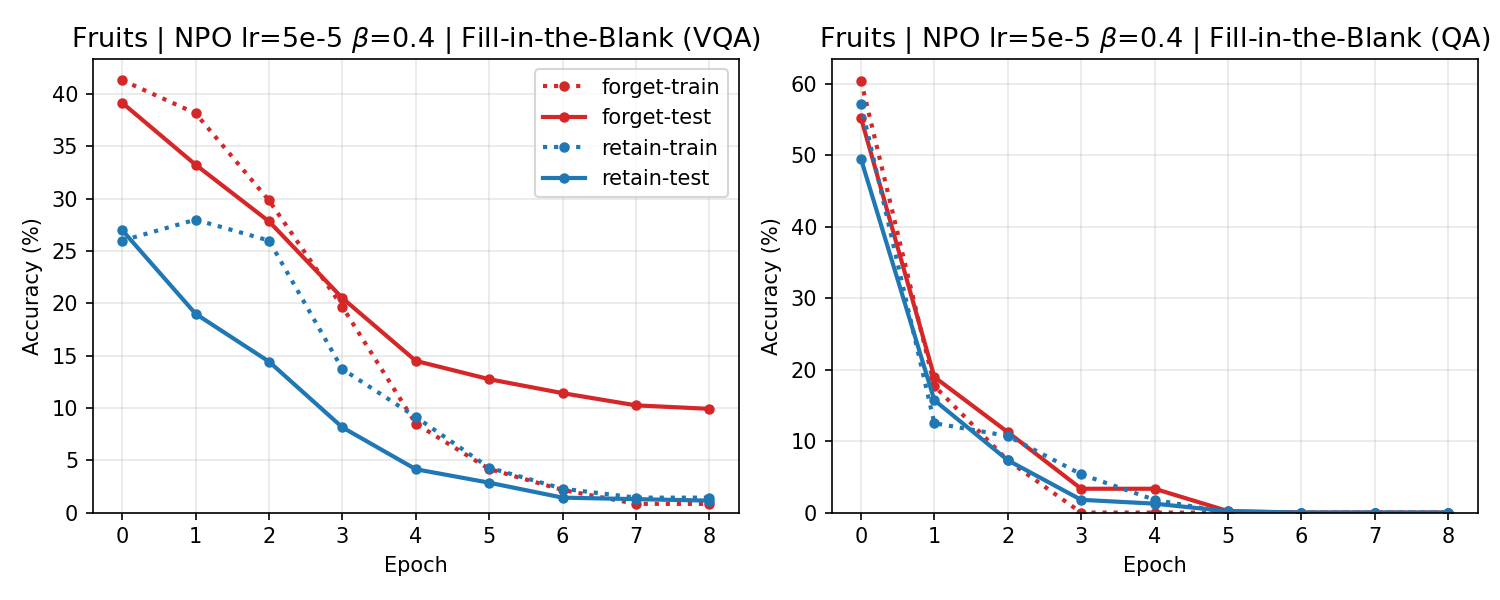}
    \includegraphics[width=0.47\linewidth]{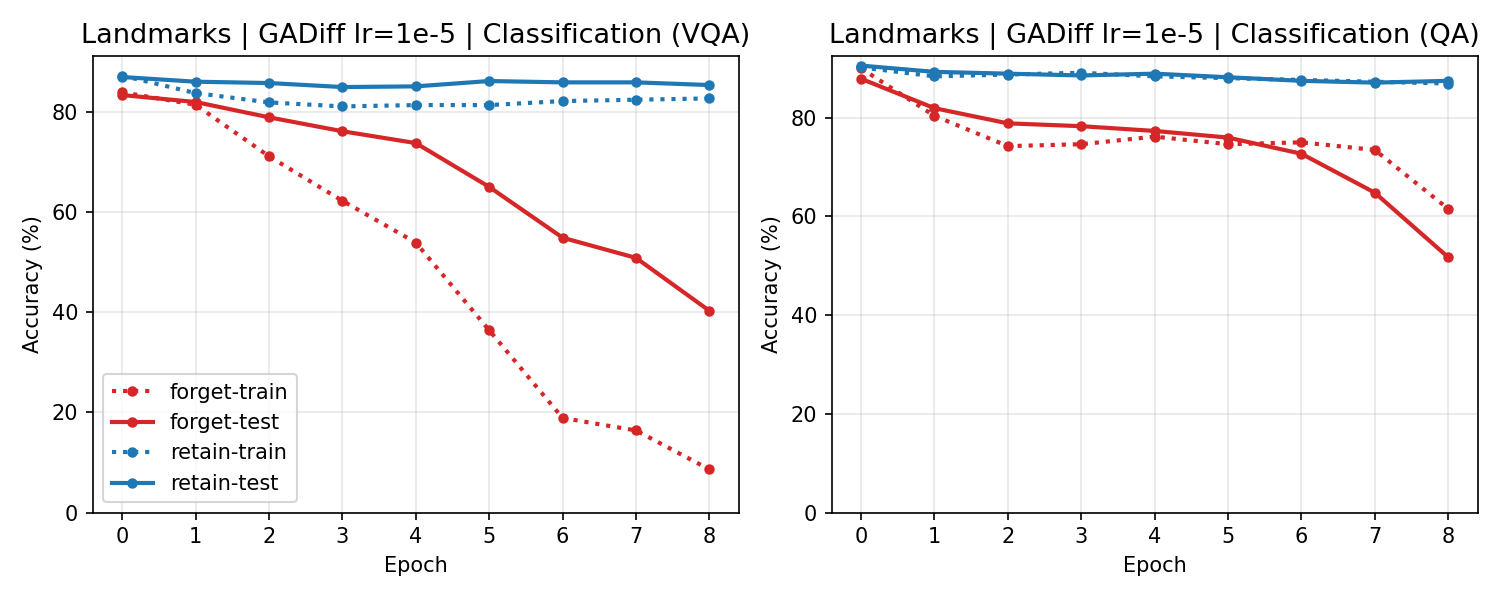}
    \caption{Comparison of unlearning generalization between image-text (left) and text-only (right) inputs. Image-text unlearning exhibits a substantially larger generalization gap than text-only unlearning, despite using the same dataset and training configurations.}
    \label{fig:image_text_vs_text_only}
\end{figure*}

\section{Pitfalls in Past Evaluations}
\label{sec:pitfalls_past_evaluations}
We discuss three major pitfalls in existing multimodal unlearning benchmarks: the lack of proper train-test splits for concept-wise unlearning, the reliance on synthetic, fine-tuned knowledge, and unfair comparisons arising from applying identical hyperparameters across different unlearning methods. All experiments are conducted on our proposed \gemmu dataset, with construction details provided in Sec.~\ref{sec:gemmu}.

\subsection{Testing on the Training Set Is Problematic for Concept-wise Unlearning}\label{sec:train_test_split}
A common practice in prior multimodal unlearning evaluations~\citep{liu2025protecting,huo-etal-2025-mmunlearner,dontsov2025clear} is to evaluate forgetting on the same $\mathcal{D}_{\text{forget}}$ used during unlearning, and similarly evaluate retention on the same $\mathcal{D}_{\text{retain}}$. Unfortunately, this violates one of the most fundamental principles of machine learning: evaluating generalization on unseen data. A model may appear to forget the target knowledge and retain the rest, while merely overfitting to the examples seen during training. Fig.~\ref{fig:generalization gap} shows a representative example of overly optimistic training-set evaluation, where accuracy is lower on the forget set and higher on the retain set compared to their respective test sets.

One might argue that some prior works additionally evaluate general capabilities on heldout, real-world benchmarks. However, such evaluations do not address the fundamental issue. Without proper train-test splits for both $\mathcal{D}_{\text{forget}}$ and $\mathcal{D}_{\text{retain}}$, the reported forgetting and retention scores become unreliable indicators of unlearning effectiveness, despite being the primary metrics.

\paragraph{Why is this issue more important in multimodal unlearning?}In some text-only unlearning scenarios, such as forgetting specific credit card numbers or passwords, the target is an exact string, making training-set evaluation potentially adequate for such \textbf{instance-wise unlearning}. In contrast, the same visual concept (e.g., a photo of Tokyo Tower) can appear in images with different viewpoints, backgrounds, and styles. This introduces greater variation in the input space, making generalization particularly important for \textbf{concept-wise unlearning}. Fig.~\ref{fig:image_text_vs_text_only} shows examples where image-text unlearning (concepts provided in images) exhibits a substantially larger generalization gap than text-only unlearning (concepts provided in pure text), highlighting the risk of overestimating forgetting in multimodal machine unlearning when evaluating only on training examples.

\subsection{Reliance on Synthetic, Fine-tuned Knowledge}
\label{sec:reliance}
Another key limitation of existing multimodal unlearning benchmarks is their reliance on knowledge introduced through fine-tuning, often in the form of synthetic facts or identities. While such settings offer a controlled environment for evaluating unlearning, they may not accurately reflect the challenges of real-world knowledge removal. In particular, unlearning fine-tuned knowledge and pre-trained knowledge may differ substantially in their underlying dynamics and difficulty.

Synthetic knowledge introduced through fine-tuning often consists of relatively isolated facts that are less entangled with the model's existing knowledge. \textit{Reverse fine-tuning}, which underlies many existing unlearning algorithms, may provide a relatively straightforward way to undo the effects of fine-tuning and thereby achieve unlearning. In contrast, real-world, pre-trained knowledge is acquired through extensive pre-training and may be deeply entangled with other useful knowledge, making its removal more challenging. As a result, strong performance on synthetic or fine-tuned knowledge does not necessarily translate to effective unlearning of real-world, pre-trained knowledge.

To better understand the differences between unlearning pre-trained and fine-tuned knowledge, we investigate both settings from two perspectives: (1) learning dynamics and (2) the forgetting–retention trade-off. These two perspectives provide complementary empirical and theoretical insights into unlearning difficulty.

\begin{figure*}
    \centering
    \includegraphics[width=0.85\linewidth]{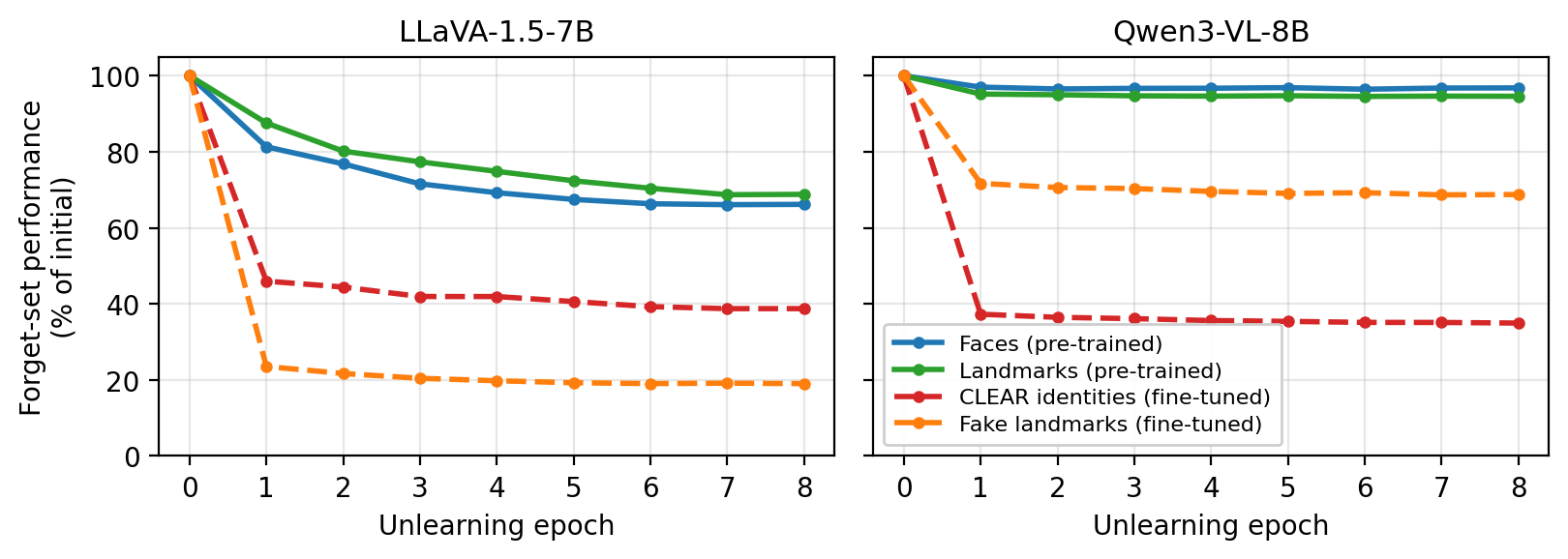}
    \caption{Forget-set image-textual performance (relative to the model before unlearning) over NPO unlearning epochs under identical hyperparameters, on LLaVA-1.5-7B (left) and Qwen3-VL-8B-Instruct (right). Fine-tuned knowledge (dashed) collapses much more quickly than pre-trained knowledge (solid).}
    \label{fig:npo_dynamics}
\end{figure*}

\paragraph{Learning Dynamics.} We compare the two settings
directly by tracking forget-set performance over
unlearning epochs on two matched pairs of GEMMU subsets: real faces versus synthetic identities, and real landmarks versus fictitious ones, where the second subset of each pair is injected into the model through fine-tuning.

In Fig.~\ref{fig:npo_dynamics}, fine-tuned knowledge drops sharply within the first epoch, whereas pre-trained knowledge decays much more slowly on both LLaVA-1.5-7B and Qwen3-VL-8B-Instruct. Plots for other unlearning algorithms are provided in Appendix~\ref{appx:learning_dynamics}.

\paragraph{Forgetting–Retention Trade-off.}
Another way to compare unlearning fine-tuned and pre-trained knowledge is to examine the retention cost required to achieve a given level of forgetting. To formalize this intuition, we present Proposition ~\ref{prop:forget-retain}.

\paragraph{Definitions.} Let $s_{\mathrm{target}}(\theta)$ and $s_{\mathrm{heldout}}(\theta)$ denote knowledge scores for the target and heldout concepts, respectively, with higher values indicating better performance. We say that forgetting is successful when $s_{\mathrm{target}}$ is below a certain threshold $\tau$, and define the required amount of forgetting as $\gamma=s_\mathrm{target}(\theta_0)-\tau>0$. The retention cost is then
$s_{\mathrm{heldout}}(\theta_0)-s_{\mathrm{heldout}}(\theta_u)$.

\begin{proposition}[Forgetting--Retention Trade-off]
\label{prop:forget-retain}
\leavevmode\par\vspace{-2em}
\begin{enumerate}
    \setlength{\itemsep}{0pt}
    \setlength{\parsep}{0pt}
    \item[I.] For fine-tuned knowledge, zero retention cost is attainable by reverting to the original model parameters $\theta_0$.
    \item[II.] For pre-trained knowledge,
    an unlearned model $\theta_u$ satisfying the forgetting
    criterion and assumptions incurs a strictly positive
    retention cost:
    \[
        s_{\mathrm{heldout}}(\theta_0)
        - s_{\mathrm{heldout}}(\theta_u)
        \geq
        \frac{\mu\gamma^2}{2K^2}
        - \frac{\varepsilon_0\gamma}{K}
        > 0.
    \]
    under the assumptions specified in Appendix~\ref{appx:assumptions}. Here, $K$, $\mu$, and $\varepsilon_0$ are constants
    defined in Appendices~\ref{appx:setting}
    and~\ref{appx:assumptions}.
\end{enumerate}
\end{proposition}

The proof is provided in Appendix~\ref{appx:proof_of_proposition}. This result suggests that unlearning pre-trained knowledge can be viewed as more challenging than unlearning fine-tuned knowledge, as successful forgetting necessarily comes at the cost of other useful knowledge under our assumptions.

\begin{figure}[ht]
    \centering
    \includegraphics[width=0.85\linewidth]{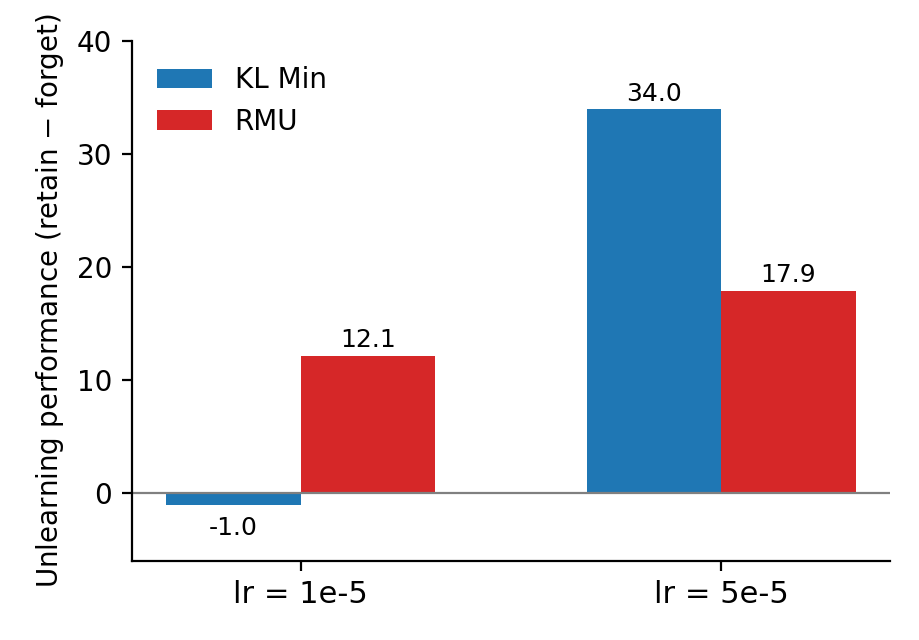}
    \caption{KL Min and RMU under two shared learning rates on
    LLaVA-1.5-7B. The winner flips with the shared configuration,
    showing that one-size-fits-all comparisons are unreliable.}
    \label{fig:hp_flip}
\end{figure}
\begin{figure}[ht]
    \centering
    \includegraphics[width=\linewidth]{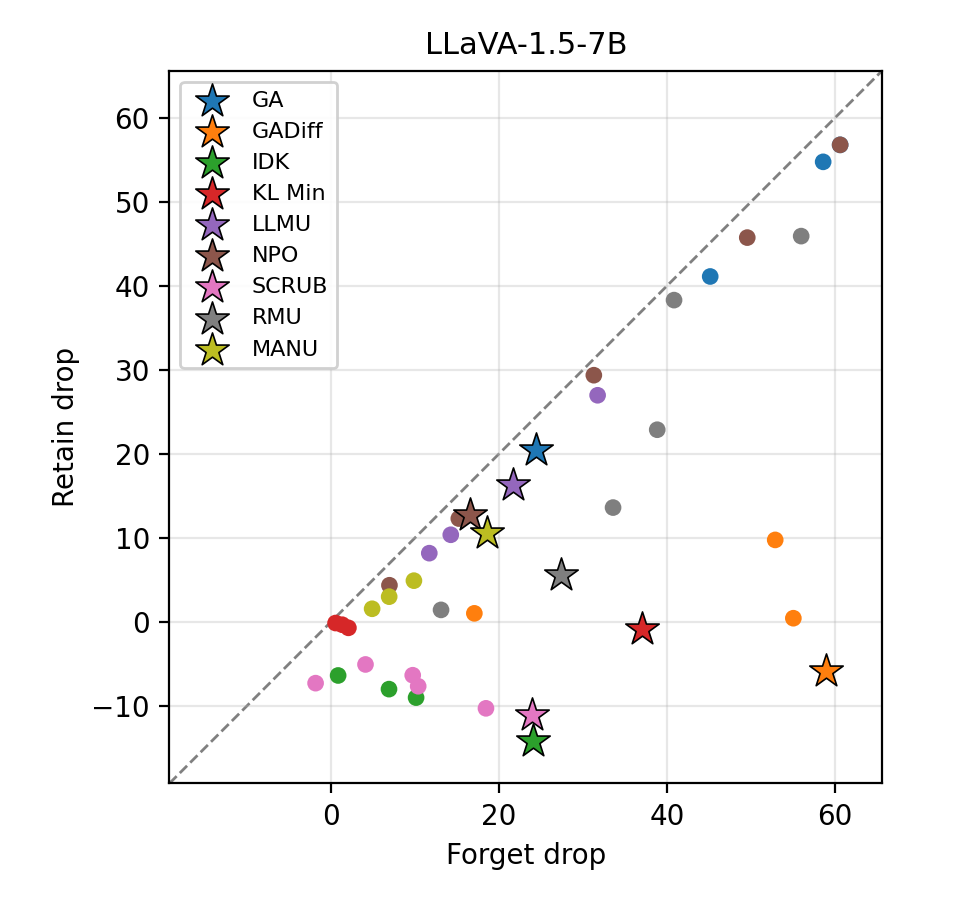}
    \caption{Forget-test drop versus retain-test drop for
    every configuration in our search space on LLaVA-1.5-7B. Without proper tuning, many configurations lie on the $y=x$ line, where retained knowledge is damaged as much as the target
    is forgotten.}
    \label{fig:scatter}
\end{figure}

\begin{table*}[t]
    \centering
    \resizebox{\linewidth}{!}{%
    \begin{tabular}{lccccccccccccc}
        \toprule
        \multirow{3}{*}{\textbf{Methods}}
        & \multicolumn{6}{c}{Forget Test ($\downarrow$)}
        & \multicolumn{6}{c}{Retain Test ($\uparrow$)}
        & \multirow{3}{*}{\textbf{Rank}} \\
        \cmidrule(lr){2-7} \cmidrule(lr){8-13}
        & \multicolumn{2}{c}{Classification} & \multicolumn{2}{c}{FITB} & \multicolumn{2}{c}{Generation}
        & \multicolumn{2}{c}{Classification} & \multicolumn{2}{c}{FITB} & \multicolumn{2}{c}{Generation} & \\
        \cmidrule(lr){2-3} \cmidrule(lr){4-5} \cmidrule(lr){6-7}
        \cmidrule(lr){8-9} \cmidrule(lr){10-11} \cmidrule(lr){12-13}
        & VQA & QA & VQA & QA & VQA & QA
        & VQA & QA & VQA & QA & VQA & QA & \\
        \midrule
        \multicolumn{14}{c}{\textit{LLaVA-1.5-7B}} \\
        \midrule
        \textit{Pre-trained} & 74.86 & 79.89 & 51.59 & 70.92 & 15.77 & 70.47 & 72.29 & 80.00 & 40.38 & 65.79 & 13.26 & 68.81 & -- \\
        GA {\scriptsize \citep{thudi2022unrolling}}      & 64.76 & 65.87 & 20.79 & 39.06 &  7.70 & 19.28 & 60.97 & 66.37 & 18.97 & 44.93 &  6.95 & 19.45 & 8 \\
        GADiff {\scriptsize \citep{liu2022continual}}    &  \textbf{4.33} &  \textbf{0.00} &  \underline{2.55} &  \textbf{0.00} &  \underline{3.20} &  \textbf{0.00} & 48.49 & 69.29 & 54.61 & \underline{87.52} & \textbf{34.18} & \textbf{81.50} & 1 \\
        SCRUB {\scriptsize \citep{kurmanji2023towards}}  & 19.17 & 75.19 &  9.27 & 51.90 &  3.87 & 60.48 & \underline{69.50} & \textbf{80.03} & \underline{59.04} & \textbf{87.81} & 29.52 & \underline{80.75} & 4 \\
        KL Min {\scriptsize \citep{maini2024tofu}}       & 47.26 & \underline{30.81} & 23.81 & \underline{14.52} &  7.13 & \underline{18.24} & 68.87 & \underline{79.31} & 41.87 & 70.96 & 14.28 & 70.42 & 3 \\
        NPO {\scriptsize \citep{zhang2024negative}}      & 72.08 & 79.34 & 33.57 & 43.40 & 13.31 & 22.83 & 69.18 & 77.93 & 30.12 & 51.05 & 12.49 & 23.45 & 9 \\
        IDK {\scriptsize \citep{maini2024tofu}}          & 60.81 & 71.82 &  6.10 & 34.81 &  3.30 & 42.52 & \textbf{69.69} & 76.26 & \textbf{87.80} & 85.45 & \underline{33.85} & 72.82 & 2 \\
        LLMU {\scriptsize \citep{yao2024large}}          & 66.84 & 73.50 & 23.85 & 24.02 &  5.96 & 39.10 & 65.06 & 77.05 & 21.59 & 29.97 &  6.94 & 42.53 & 7 \\
        RMU {\scriptsize \citep{li2024wmdp}}             &  \underline{4.54} & 74.11 &  \textbf{1.16} & 64.13 &  \textbf{1.06} & 54.63 & 65.74 & 75.86 & 34.34 & 58.02 & 18.21 & 54.88 & 5 \\
        MANU {\scriptsize \citep{liu2025modality}}       & 60.14 & 70.35 & 20.68 & 37.52 &  5.11 & 58.41 & 63.78 & 74.75 & 25.99 & 43.55 &  7.29 & 61.81 & 6 \\
        \midrule
        \multicolumn{14}{c}{\textit{Qwen3-VL-8B-Instruct}} \\
        \midrule
        \textit{Pre-trained} & 88.82 & 86.32 & 82.61 & 78.37 & 42.11 & 59.49 & 86.69 & 85.25 & 76.39 & 78.20 & 43.16 & 59.23 & -- \\
        GA {\scriptsize \citep{thudi2022unrolling}}      & 67.76 & 65.97 & 34.62 & 38.45 &  3.76 &  \underline{8.53} & 64.83 & 67.90 & 28.28 & 37.46 &  4.76 & 12.41 & 8 \\
        GADiff {\scriptsize \citep{liu2022continual}}    & 13.91 & \textbf{31.10} &  \textbf{0.10} &  \textbf{0.00} &  \textbf{0.23} &  \textbf{0.00} & 56.70 & 68.09 & 58.62 & 57.46 & \textbf{53.01} & \underline{73.35} & 2 \\
        SCRUB {\scriptsize \citep{kurmanji2023towards}}  &  \underline{3.07} & 41.22 &  \underline{1.39} & 31.34 &  \underline{0.80} & 20.56 & 84.11 & \underline{87.04} & 76.03 & \textbf{86.24} & \underline{44.72} & \textbf{73.97} & 1 \\
        KL Min {\scriptsize \citep{maini2024tofu}}       & 88.81 & 86.10 & 79.60 & 73.65 & 40.11 & 57.38 & \textbf{86.76} & 85.12 & \underline{77.41} & \underline{78.31} & 42.20 & 59.41 & 7 \\
        NPO {\scriptsize \citep{zhang2024negative}}      & 68.63 & 66.17 & 22.06 & 43.92 &  3.10 & 10.58 & 65.44 & 68.37 & 19.48 & 40.31 &  2.43 & 11.26 & 9 \\
        IDK {\scriptsize \citep{maini2024tofu}}          & 65.17 & 86.38 &  7.83 & 38.88 &  2.66 & 13.73 & 53.84 & \textbf{90.43} & \textbf{86.92} & 67.09 & 42.53 & 62.25 & 3 \\
        LLMU {\scriptsize \citep{yao2024large}}          & 86.17 & 83.76 & 34.80 & 42.65 & 19.70 & 25.73 & \underline{84.78} & 83.94 & 35.32 & 49.97 & 18.56 & 27.38 & 6 \\
        RMU {\scriptsize \citep{li2024wmdp}}             &  \textbf{2.48} & 83.91 &  1.66 & 76.88 &  4.02 & 59.22 & 84.34 & 85.28 & 71.44 & 75.18 & 41.01 & 58.98 & 4 \\
        MANU {\scriptsize \citep{liu2025modality}}       & 49.65 & \underline{35.62} & 21.37 & \underline{20.26} & 34.68 & 41.02 & 76.31 & 65.08 & 48.80 & 46.02 & 40.64 & 55.26 & 5 \\
        \bottomrule
    \end{tabular}%
    }
    \caption{Benchmark results on LLaVA-1.5-7B and Qwen3-VL-8B-Instruct, averaged over the four real-world subsets. VQA denotes image-textual and QA pure-text evaluation. \textbf{Bold} and \underline{underline} denote the best and second-best result per column within each model; Rank orders methods by average retain-test minus average forget-test performance. The classic and simple GADiff achieves the strongest overall balance between forgetting and retention.}
    \label{tab:main_results}
\end{table*}

\subsection{Hyperparameter Sensitivity}\label{sec:param_sen}
Most existing benchmarks compare unlearning methods under a single shared training configuration (Tab.~\ref{tab:benchmarks_comparisons}).
This implicitly assumes that different methods share a common
operating range, which we find to be far from true. \footnote{For benchmarks that do not report hyperparameter selection in the paper, we classify them by the hyperparameters in their released code.}

Fig.~\ref{fig:hp_flip} compares KL Min and RMU trained under
identical shared configurations on LLaVA-1.5-7B, measured by average retain-set minus average forget-set performance on heldout examples, where higher means better forgetting at lower retention cost. At a learning rate of $10^{-5}$, RMU wins clearly while KL Min appears entirely ineffective; at $5\times10^{-5}$, the conclusion reverses, with KL Min now outperforming RMU by a wide margin. The same shared
protocol thus supports two contradictory conclusions about which algorithm is better.

Fig.~\ref{fig:scatter} extends this observation to all nine methods across our search space. Without proper tuning, many configurations lie on the $y=x$ line, i.e., indiscriminate
forgetting. After per-method search, the strongest methods fall well below it. This sensitivity partly reflects that multimodal unlearning is harder than text-only unlearning, where a shared configuration is common practice \cite{maini2024tofu}, and further indicates that fair evaluation for multimodal unlearning requires a per-method hyperparameter search. More examples are provided in Appendix~\ref{appx:hyperparameter_sensitivity}. 

\section{\gemmu}
\label{sec:gemmu}
\gemmu is designed to reveal the pitfalls discussed in Sec.~\ref{sec:pitfalls_past_evaluations} and to provide practical guidelines that future evaluation benchmarks should follow. Its core components include (1) concept removal of existing real-world knowledge and comparison with fine-tuned knowledge; (2) generalization of forgetting and retention to heldout examples; (3) per-method hyperparameter search for fair comparisons; and additionally, (4) modality gap analysis using image-text and text-only questions, inspired by~\citet{wang2025umu} and ~\citet{huo-etal-2025-mmunlearner}. We first describe our methodology (Sec.~\ref{sec:construction}), followed by a re-benchmarking of existing unlearning methods (Sec.~\ref{sec:rebenchmarking}), and conclude with design principles for future benchmarks (Sec.~\ref{sec:principles}).

\subsection{Benchmark Construction}
\label{sec:construction}
\paragraph{Concept removal for real-world and synthetic knowledge.}
To achieve a diverse source of real-world knowledge, we consider human identities, landmarks, car brands, and fruits as our main categories, with 139 concepts in total. Following ~\citet{liu2025protecting} and ~\citet{wang2025umu}, 
each concept is assigned exclusively to the forget or retain set. We also verified that the evaluated MLLMs have a decent understanding of those concepts before unlearning in Tab.~\ref{tab:main_results}.

To compare the unlearning of pre-trained and fine-tuned knowledge in Sec.~\ref{sec:reliance}, we additionally include two synthetic knowledge subsets: CLEAR and fake landmarks. For CLEAR, we adopt fictitious identities, along with their associated images and biographical information from the CLEAR dataset \citep{dontsov2025clear}. For fake landmarks, we generate fictional landmark profiles with corresponding images and construct questions about their attributes, including location and completion date. Each synthetic subset consists of 40 concepts.

\begin{table}[t]
    \centering
    \small
    \resizebox{\linewidth}{!}{%
    \begin{tabular}{lccccl}
        \toprule
         \multirow{3}{*}{\textbf{Methods}}& \multicolumn{2}{c}{LLaVA-1.5-7B} & \multicolumn{2}{c}{Qwen3-VL-8B} & \multirow{3}{*}{\textbf{Removal type}}\\
        \cmidrule(lr){2-3} \cmidrule(lr){4-5}
        & \makecell{VQA \\drop (\%)} & \makecell{QA \\drop (\%)} & \makecell{VQA \\drop (\%)} & \makecell{QA \\drop (\%)} &  \\
        \midrule
        GA     & 34.4 & 43.9 & 50.3 & 49.6 & Whole concept \\
        GADiff & 92.9 & 100.0 & 93.3 & 86.1 & Whole concept \\
        SCRUB  & 77.3 & 15.2 & 97.5 & 58.5 & Association \\
        KL Min & 45.0 & 71.3 &  2.4 &  3.1 & Inconsistent \\
        NPO    & 16.4 & 34.2 & 56.1 & 46.2 & Whole concept \\
        IDK    & 50.6 & 32.6 & 64.6 & 38.0 & Association \\
        LLMU   & 32.0 & 38.3 & 34.1 & 32.1 & Whole concept \\
        RMU    & 95.3 & 12.8 & 96.2 &  1.9 & Association \\
        MANU   & 39.6 & 24.9 & 50.5 & 56.8 & Whole concept \\
        \bottomrule
    \end{tabular}%
    }
    \caption{Forget-test performance drop relative to the pre-trained
    model (\% of initial), per modality, averaged over the four
    real-world subsets at each method's selected configuration.
    Association removers suppress VQA while QA largely survives while
    whole-concept removers suppress both.}
    \label{tab:removal_type}
\end{table}
\paragraph{Generalizability analysis.} Within each concept, examples are split into disjoint training and test sets, so the test sets assess unseen examples of the same concepts. Images are collected from the Internet and manually verified by the authors.\footnote{We will release the source links instead of the images to prevent copyright infringement.}  We adopt a train-test split of 5:10, which is less extreme than the single-training-data setup in MMUBench \cite{li2024single}.

\paragraph{Per-method hyperparameter search.}
For each method, we search for its optimal hyperparameters by scoring each configuration by its average retain-train minus average forget-train performance and select the configuration with the best score averaged over the four real-world subsets. Selection uses training splits only so that test data never influences it. See Appendix~\ref{appx:implementation_details} for the selected
hyperparameters and implementation details.

\paragraph{Modality gap analysis.} Following  \citet{liu2025protecting} and \citet{dontsov2025clear}, we construct questions in two input modalities: visual question answering (VQA) pairs an image with a textual question, while text-only question answering (QA) contains the concept name and no image. Both modalities cover single-choice classification, fill-in-the-blank, and generation tasks. We use accuracy for classification and fill-in-the-blank tasks and ROUGE-L for generation. For both image-text and text-only tasks, we paraphrase the question texts while preserving their meaning and ground-truth answers to prevent overfitting to the question formats. 

Representative examples from the forget and retain training and test sets of all six subsets are provided in Appendix~\ref{appx:representative_examples}.

\subsection{Participating MLLMs}
We consider Qwen3-VL-8B-Instruct~\citep{bai2025qwen3} and LLaVA-1.5-7B~\citep{liu2024improved} as our base MLLMs. Following prior works on multimodal unlearning~\citep{huo-etal-2025-mmunlearner,dontsov2025clear,zheng2026offside}, we adopt LoRA~\citep{hu2022lora} for both fine-tuning the base models in the synthetic versus pre-trained knowledge experiment and applying the unlearning algorithms.

\subsection{Machine Unlearning Methods}
We evaluate nine representative machine unlearning methods covering a diverse range of unlearning strategies. Four are built on \emph{gradient ascent} over forget-set responses: GA~\citep{thudi2022unrolling} uses the ascent loss alone;
GADiff~\citep{liu2022continual} adds gradient descent on the retain set; KL Min~\citep{maini2024tofu} instead regularizes retain updates with a KL penalty toward the pre-trained model; and LLMU~\citep{yao2024large} further pairs forget prompts with unrelated responses under a KL penalty. NPO~\citep{zhang2024negative} replaces unbounded ascent with a preference-style loss whose ascent weight shrinks as the forget loss exceeds the pre-trained loss.
IDK~\citep{maini2024tofu} simply relabels forget answers with abstention targets. SCRUB~\citep{kurmanji2023towards} is \emph{distillation-based}, maximizing KL divergence from a frozen
teacher on forget examples while minimizing it on retain examples. Finally, two methods operate on \emph{internal representations}: RMU~\citep{li2024wmdp} steers forget-set activations at a chosen layer toward a random vector while anchoring retain-set activations, and MANU~\citep{liu2025modality} prunes MLP neurons ranked by their forget-versus-retain importance.

\subsection{Re-benchmarking Unlearning Algorithms}\label{sec:rebenchmarking}
Table~\ref{tab:main_results} shows results for nine unlearning methods under \textbf{GEMMU}. Our main findings are as follows.

\paragraph{Simple methods win.} Recent methods are often reported as state-of-the-art on prior benchmarks. For example, MANU \cite{liu2025modality} outperforms GADiff on MLLMU-Bench \cite{liu2025protecting}. Rankings on \gemmu differ: the two strongest methods are among the oldest and simplest, GADiff and SCRUB.

\paragraph{Association removal vs.\ concept removal.}
\label{sec:modality_gap}
The two modalities reveal \emph{what} is unlearned. Following the notion in \citet{huo-etal-2025-mmunlearner}, if only VQA drops while QA survives, the model has merely severed the association between
visual appearance and knowledge; if both drop, the concept itself is removed. Tab.~\ref{tab:removal_type} classifies each method by comparing its VQA and QA drops on the forget-test split based on the results in Tab.~\ref{tab:main_results}. RMU is the clearest association remover: on Qwen3-VL-8B, VQA classification falls to 2.48 while QA stays at 83.91 (FITB: 1.66 vs.\ 76.88), so the model no longer recognizes the visual entity but still knows everything about it. GADiff removes whole concepts, at the cost of the largest retain-side drop.
\paragraph{No method solves the task.}
Even the best entries in Table~\ref{tab:main_results} fall short of ideal unlearning: GADiff sacrifices retain classification (48.49 on LLaVA, 56.7 on Qwen), SCRUB leaves textual forget knowledge largely intact, and the remaining methods either under-forgets or over-destroys. We conclude that unlearning real-world, pre-trained multimodal knowledge remains unsolved.

\subsection{On Guidelines for Future Multimodal Unlearning Benchmarks}\label{sec:principles}
We close by distilling our findings into design guidance, in the hope
that good practices are inherited by future benchmarks.

\paragraph{What makes a good design?} First, the fundamentals of machine
learning evaluation: a proper train--test split, so that forgetting and
retention are measured as generalization rather than memorization
(Sec.~\ref{sec:train_test_split}), and a hyperparameter search per
method, without which conclusions about relative effectiveness may reflect the chosen configuration rather than the methods themselves (Sec.~\ref{sec:param_sen}). Second, echoing modality-gap
studies in multimodal
unlearning~\citep{wang2025umu,huo-etal-2025-mmunlearner}, benchmarks
should construct and report both VQA and QA metrics for every task.
Only the contrast between the two reveals whether a method removes the
visual--textual association or the concept itself
(Sec.~\ref{sec:rebenchmarking}).

\paragraph{What to avoid?} Inheriting text-only practices without
accounting for what is unique to MLLMs: instance-wise forgetting may be
defensible for text, but not for concept-wise forgetting in MLLMs
(Fig.~\ref{fig:image_text_vs_text_only}). Likewise, convenience should
not trump rigor. Benchmarks built on synthetic, fine-tuned knowledge
are easier to construct but may not reflect realistic unlearning
(Sec.~\ref{sec:reliance}), and applying one shared hyperparameter
configuration to every method can distort the comparison
(Sec.~\ref{sec:param_sen}).

\section{Conclusion}
In this work, we introduce \textbf{GEMMU}, a benchmark for multimodal machine unlearning that emphasizes generalization-aware evaluation and covers a broader and more diverse range of real-world concepts than prior benchmarks. Experiments on \textbf{GEMMU} reveal substantial generalization gaps overlooked by previous benchmarks, highlight the challenges of removing pre-trained knowledge that is deeply entangled with retained knowledge, and demonstrate how proper hyperparameter tuning can alter the relative rankings of unlearning methods.

As unlearning methods and benchmarks become increasingly sophisticated, we emphasize that fundamental principles of evaluation must not be overlooked to ensure that scientific progress is built on a solid foundation.
\section*{Limitations}
Following common practice in existing multimodal unlearning work, we use LoRA \citep{hu2022lora} for both fine-tuning and unlearning. Full-model fine-tuning may exhibit different unlearning behavior from adapter-based fine-tuning. However, we do not extend our experiments to full-model fine-tuning due to its substantially higher computational cost, and instead focus on settings that can be feasibly evaluated on an NVIDIA H100 GPU. 
\bibliography{custom}

@inproceedings{dontsov2025clear,
    title = "{CLEAR}: Character Unlearning in Textual and Visual Modalities",
    author = "Dontsov, Alexey  and
      Korzh, Dmitrii  and
      Zhavoronkin, Alexey  and
      Mikheev, Boris  and
      Bobkov, Denis  and
      Alanov, Aibek  and
      Rogov, Oleg  and
      Oseledets, Ivan  and
      Tutubalina, Elena",
    editor = "Che, Wanxiang  and
      Nabende, Joyce  and
      Shutova, Ekaterina  and
      Pilehvar, Mohammad Taher",
    booktitle = "Findings of the Association for Computational Linguistics: ACL 2025",
    month = jul,
    year = "2025",
    address = "Vienna, Austria",
    publisher = "Association for Computational Linguistics",
    url = "https://aclanthology.org/2025.findings-acl.1058/",
    doi = "10.18653/v1/2025.findings-acl.1058",
    pages = "20582--20603",
    ISBN = "979-8-89176-256-5"
}

@inproceedings{huo-etal-2025-mmunlearner,
    title = "{MMU}nlearner: Reformulating Multimodal Machine Unlearning in the Era of Multimodal Large Language Models",
    author = "Huo, Jiahao  and
      Yan, Yibo  and
      Zheng, Xu  and
      Lyu, Yuanhuiyi  and
      Zou, Xin  and
      Wei, Zhihua  and
      Hu, Xuming",
    editor = "Che, Wanxiang  and
      Nabende, Joyce  and
      Shutova, Ekaterina  and
      Pilehvar, Mohammad Taher",
    booktitle = "Findings of the Association for Computational Linguistics: ACL 2025",
    month = jul,
    year = "2025",
    address = "Vienna, Austria",
    publisher = "Association for Computational Linguistics",
    url = "https://aclanthology.org/2025.findings-acl.375/",
    doi = "10.18653/v1/2025.findings-acl.375",
    pages = "7190--7206",
    ISBN = "979-8-89176-256-5"
}

@article{kawakami2025pulse,
  title={Pulse: Practical evaluation scenarios for large multimodal model unlearning},
  author={Kawakami, Tatsuki and Egashira, Kazuki and Miyai, Atsuyuki and Irie, Go and Aizawa, Kiyoharu},
  journal={arXiv preprint arXiv:2507.01271},
  year={2025}
}

@inproceedings{liu2025protecting,
  title={Protecting privacy in multimodal large language models with mllmu-bench},
  author={Liu, Zheyuan and Dou, Guangyao and Jia, Mengzhao and Tan, Zhaoxuan and Zeng, Qingkai and Yuan, Yongle and Jiang, Meng},
  booktitle={Proceedings of the 2025 Conference of the Nations of the Americas Chapter of the Association for Computational Linguistics: Human Language Technologies (Volume 1: Long Papers)},
  pages={4105--4135},
  year={2025}
}

@article{bai2025qwen3,
  title={Qwen3-vl technical report},
  author={Bai, Shuai and Cai, Yuxuan and Chen, Ruizhe and Chen, Keqin and Chen, Xionghui and Cheng, Zesen and Deng, Lianghao and Ding, Wei and Gao, Chang and Ge, Chunjiang and others},
  journal={arXiv preprint arXiv:2511.21631},
  year={2025}
}

@inproceedings{
li2025llavanextinterleave,
title={{LL}a{VA}-Ne{XT}-Interleave: Tackling Multi-image, Video, and 3D in Large Multimodal Models},
author={Feng Li and Renrui Zhang and Hao Zhang and Yuanhan Zhang and Bo Li and Wei Li and Zejun MA and Chunyuan Li},
booktitle={The Thirteenth International Conference on Learning Representations},
year={2025},
url={https://openreview.net/forum?id=oSQiao9GqB}
}

@inproceedings{wang2025umu,
  title={UMU-Bench: Closing the Modality Gap in Multimodal Unlearning Evaluation},
  author={Wang, Chengye and Li, Yuyuan and Feng, XiaoHua and Chen, Chaochao and Zheng, Xiaolin and Yin, Jianwei},
  booktitle={The Thirty-ninth Annual Conference on Neural Information Processing Systems Datasets and Benchmarks Track},
  year={2025}
}

@article{xu2025pebench,
  title={Pebench: A fictitious dataset to benchmark machine unlearning for multimodal large language models},
  author={Xu, Zhaopan and Zhou, Pengfei and Tang, Weidong and Ai, Jiaxin and Zhao, Wangbo and Wang, Kai and Peng, Xiaojiang and Shao, Wenqi and Yao, Hongxun and Zhang, Kaipeng},
  journal={arXiv preprint arXiv:2503.12545},
  year={2025}
}

@inproceedings{ma2025benchmarking,
  title={Benchmarking vision language model unlearning via fictitious facial identity dataset},
  author={Ma, Yingzi and Wang, Jiongxiao and Wang, Fei and Ma, Siyuan and Li, Jiazhao and Pan, Jinsheng and Li, Xiujun and Huang, Furong and Sun, Lichao and Li, Bo and others},
  booktitle={International Conference on Learning Representations},
  volume={2025},
  pages={87806--87827},
  year={2025}
}

@inproceedings{
maini2024tofu,
title={{TOFU}: A Task of Fictitious Unlearning for {LLM}s},
author={Pratyush Maini and Zhili Feng and Avi Schwarzschild and Zachary Chase Lipton and J Zico Kolter},
booktitle={First Conference on Language Modeling},
year={2024},
url={https://openreview.net/forum?id=B41hNBoWLo}
}

@inproceedings{li2024single,
 author = {Li, Jiaqi and Wei, Qianshan and Zhang, Chuanyi and Qi, Guilin and Du, Miaozeng and Chen, Yongrui and Bi, Sheng and Liu, Fan},
 booktitle = {Advances in Neural Information Processing Systems},
 doi = {10.52202/079017-1116},
 editor = {A. Globerson and L. Mackey and D. Belgrave and A. Fan and U. Paquet and J. Tomczak and C. Zhang},
 pages = {35414--35453},
 publisher = {Curran Associates, Inc.},
 title = {Single Image Unlearning: Efficient Machine Unlearning in Multimodal Large Language Models},
 url = {https://proceedings.neurips.cc/paper_files/paper/2024/file/3e53d82a1113e3d240059a9195668edc-Paper-Conference.pdf},
 volume = {37},
 year = {2024}
}

@inproceedings{zheng2026offside,
  title={Offside: Benchmarking unlearning misinformation in multimodal large language models},
  author={Zheng, Hao and Pang, Zirui and Li, Ling and Deng, Zhijie and Pu, Yuhan and Zhu, Zhaowei and Xia, Xiaobo and Wei, Jiaheng},
  booktitle={Findings of the Association for Computational Linguistics: ACL 2026},
  pages={12602--12620},
  year={2026}
}

@inproceedings{liu2024improved,
  title={Improved baselines with visual instruction tuning},
  author={Liu, Haotian and Li, Chunyuan and Li, Yuheng and Lee, Yong Jae},
  booktitle={Proceedings of the IEEE/CVF conference on computer vision and pattern recognition},
  pages={26296--26306},
  year={2024}
}

@inproceedings{liu2025modality,
  title={Modality-aware neuron pruning for unlearning in multimodal large language models},
  author={Liu, Zheyuan and Dou, Guangyao and Yuan, Xiangchi and Zhang, Chunhui and Tan, Zhaoxuan and Jiang, Meng},
  booktitle={Proceedings of the 63rd Annual Meeting of the Association for Computational Linguistics (Volume 1: Long Papers)},
  pages={5913--5933},
  year={2025}
}

@article{kurmanji2023towards,
  title={Towards unbounded machine unlearning},
  author={Kurmanji, Meghdad and Triantafillou, Peter and Hayes, Jamie and Triantafillou, Eleni},
  journal={Advances in neural information processing systems},
  volume={36},
  pages={1957--1987},
  year={2023}
}

@article{yao2024large,
  title={Large language model unlearning},
  author={Yao, Yuanshun and Xu, Xiaojun},
  journal={Advances in Neural Information Processing Systems},
  volume={37},
  pages={105425--105475},
  year={2024}
}

@inproceedings{thudi2022unrolling,
  title={Unrolling sgd: Understanding factors influencing machine unlearning},
  author={Thudi, Anvith and Deza, Gabriel and Chandrasekaran, Varun and Papernot, Nicolas},
  booktitle={2022 IEEE 7th European symposium on security and privacy (EuroS\&P)},
  pages={303--319},
  year={2022},
  organization={IEEE}
}

@InProceedings{liu2022continual,
  title = 	 {Continual Learning and Private Unlearning},
  author =       {Liu, Bo and Liu, Qiang and Stone, Peter},
  booktitle = 	 {Proceedings of The 1st Conference on Lifelong Learning Agents},
  pages = 	 {243--254},
  year = 	 {2022},
  editor = 	 {Chandar, Sarath and Pascanu, Razvan and Precup, Doina},
  volume = 	 {199},
  series = 	 {Proceedings of Machine Learning Research},
  month = 	 {22--24 Aug},
  publisher =    {PMLR},
  url = 	 {https://proceedings.mlr.press/v199/liu22a.html}
}

@article{li2024wmdp,
  title={The wmdp benchmark: Measuring and reducing malicious use with unlearning},
  author={Li, Nathaniel and Pan, Alexander and Gopal, Anjali and Yue, Summer and Berrios, Daniel and Gatti, Alice and Li, Justin D and Dombrowski, Ann-Kathrin and Goel, Shashwat and Phan, Long and others},
  journal={arXiv preprint arXiv:2403.03218},
  year={2024}
}

@inproceedings{
zhang2024negative,
title={Negative Preference Optimization: From Catastrophic Collapse to Effective Unlearning},
author={Ruiqi Zhang and Licong Lin and Yu Bai and Song Mei},
booktitle={First Conference on Language Modeling},
year={2024},
url={https://openreview.net/forum?id=MXLBXjQkmb}
}

@inproceedings{
hu2022lora,
title={Lo{RA}: Low-Rank Adaptation of Large Language Models},
author={Edward J Hu and yelong shen and Phillip Wallis and Zeyuan Allen-Zhu and Yuanzhi Li and Shean Wang and Lu Wang and Weizhu Chen},
booktitle={International Conference on Learning Representations},
year={2022},
url={https://openreview.net/forum?id=nZeVKeeFYf9}
}
\appendix


\begin{figure*}[]
    \centering
    \includegraphics[width=0.9\linewidth]{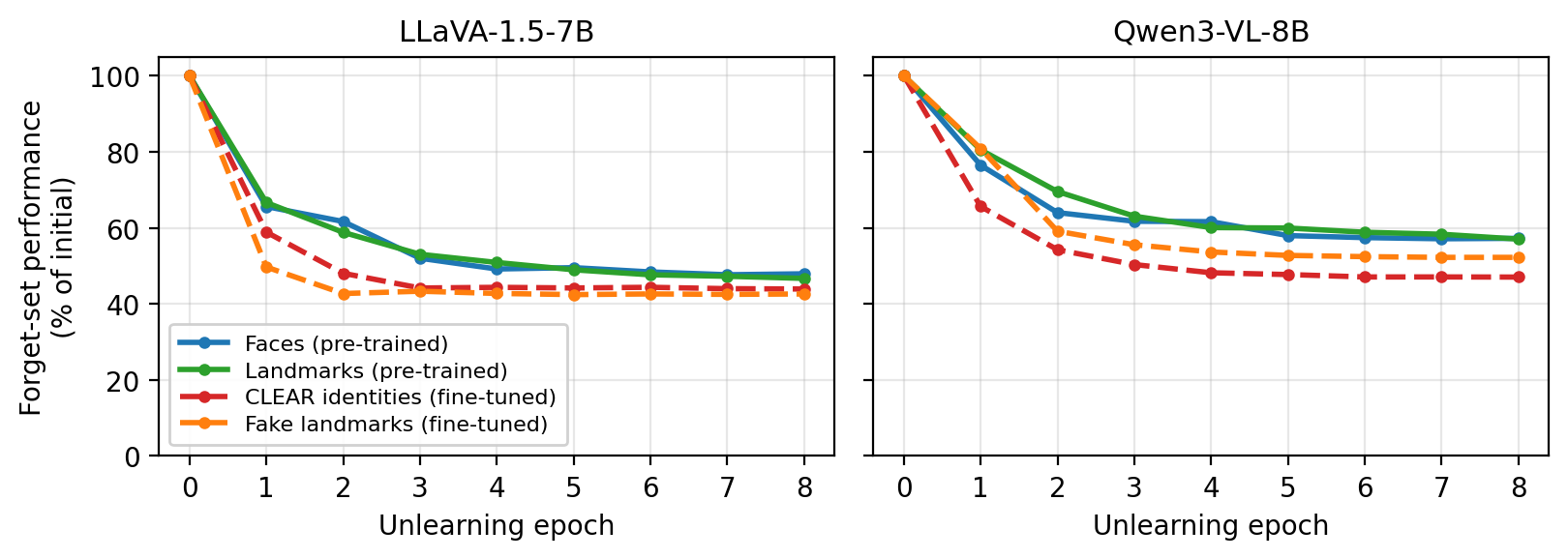}
    \caption{Forget-set image-textual performance (relative to the model before unlearning) over unlearning epochs, averaged over all 8 learning-based unlearning methods (i.e., excluding the pruning-based MANU), on LLaVA-1.5-7B (left) and Qwen3-VL-8B-Instruct (right). Fine-tuned knowledge (dashed) collapses more quickly than pre-trained knowledge (solid).}
    \label{fig:average_dynamics}
\end{figure*}

\section{Extended Results for Learning Dynamics}
\label{appx:learning_dynamics}
Fig.~\ref{fig:average_dynamics} shows the learning dynamics averaged across all eight learning-based unlearning methods. The findings in Sec.~\ref{sec:reliance} remain consistent, with pre-trained knowledge being forgotten more slowly than fine-tuned knowledge. The gap is smaller in the averaged results because some unlearning methods either fail to induce meaningful forgetting or cause the model to collapse entirely.

\section{Proof of Proposition~\ref{prop:forget-retain}}
\label{appx:proof_of_proposition}

\makeatletter
\@ifundefined{assumption}{\newtheorem{assumption}{Assumption}}{}
\makeatother
\renewcommand{\theassumption}{A\arabic{assumption}}

\subsection{Analysis setup}
\label{appx:setting}

We use differentiable knowledge scores $s_{\mathrm{target}}$ and $s_{\mathrm{heldout}}$ for the analysis, such as the average log-likelihood of reference answers. Let $\Theta$ be a neighborhood containing $\theta_0$ where the assumptions below hold, and let $\|\cdot\|_2$ denote the Euclidean norm. Consider any updated model $\theta_u\in\Theta$ satisfying the forgetting criterion $s_{\mathrm{target}}(\theta_u)\leq\tau$.

We measure retention cost relative to the original model as $s_{\mathrm{heldout}}(\theta_0)-s_{\mathrm{heldout}}(\theta_u)$. For fine-tuned knowledge, we show that $\theta_u=\theta_0$ satisfies the forgetting criterion with zero retention cost. For pre-trained knowledge, we establish a strictly positive lower bound on the retention cost under the following assumptions.

\subsection{Assumptions}
\label{appx:assumptions}

The following conditions are needed for the pre-trained case. Here, the target concept is already present in $\theta_0$, and
\[
    \gamma := s_{\mathrm{target}}(\theta_0)-\tau > 0
\]
is the score reduction required to reach the forgetting threshold.

\begin{assumption}[Lipschitz target score]
\label{as:lip}
\leavevmode\par
$s_{\mathrm{target}}$ is $K$-Lipschitz, i.e., there is $K>0$ such that, for all $\theta,\theta'\in\Theta$,
\[
    |s_{\mathrm{target}}(\theta)-s_{\mathrm{target}}(\theta')|
    \leq K\|\theta-\theta'\|_2.
\]
\end{assumption}
This condition limits how much the target score can change per unit of parameter displacement. Reducing the score by at least $\gamma$ therefore requires a displacement of at least $\gamma/K$.

\begin{assumption}[Limited local improvement]
\label{as:stat}
\leavevmode\par
At $\theta_0$, we bound how quickly retained utility can improve under small parameter changes: for some $\varepsilon_0\geq0$,
\[
    \|\nabla s_{\mathrm{heldout}}(\theta_0)\|_2\leq\varepsilon_0.
\]
\end{assumption}
This reflects the assumption that the pre-trained model is near-optimal on the heldout concepts. Even in the most favorable direction, an update $\delta$ provides at most $\varepsilon_0\|\delta\|_2$ of improvement under the linear approximation.

\begin{assumption}[Quadratic penalty]
\label{as:curv}
\leavevmode\par
There is $\mu>0$ such that, for every $\theta_0+\delta\in\Theta$,
\[
\begin{aligned}
    s_{\mathrm{heldout}}(\theta_0+\delta)
    \leq {}& s_{\mathrm{heldout}}(\theta_0)
    \\+\langle\nabla s_{\mathrm{heldout}}(\theta_0),\delta\rangle 
    &-\frac{\mu}{2}\|\delta\|_2^2.
\end{aligned}
\]
\end{assumption}
This condition formalizes the intuition that pre-trained knowledge is entangled with other useful knowledge: it places retained utility below its first-order approximation by at least $\mu\|\delta\|_2^2/2$ for every update within $\Theta$. The penalty grows quadratically with displacement, whereas the possible first-order improvement in Assumption~\ref{as:stat} grows only linearly. Together, they imply a positive retention cost when the displacement is sufficiently large.

To ensure that successful forgetting requires such a displacement, we additionally require
\begin{equation}
\label{eq:eps_condition}
    \varepsilon_0 < \frac{\mu\gamma}{2K}.
\end{equation}
Equivalently, $\gamma/K>2\varepsilon_0/\mu$: the minimum displacement needed for forgetting exceeds the distance at which the quadratic penalty outweighs the possible first-order improvement.

\subsection{Proof}
\label{appx:proof}
The two cases differ in whether $\theta_0$ already satisfies the forgetting criterion. In part~I, returning to $\theta_0$ gives a feasible solution with zero retention cost relative to that same model. In part~II, its target score exceeds the threshold by $\gamma>0$. Successful forgetting then requires moving away from $\theta_0$, and the assumptions translate that required displacement into a positive retention cost.
\paragraph{Fine-tuned knowledge (part~I).}
Let $\theta_{\mathrm{ft}}=\theta_0+\Delta_{\mathrm{ft}}$, where
\[
    s_{\mathrm{target}}(\theta_0)\leq\tau<s_{\mathrm{target}}(\theta_{\mathrm{ft}}).
\]
Reversing the fine-tuning update gives $\theta_u=\theta_{\mathrm{ft}}-\Delta_{\mathrm{ft}}=\theta_0$. This model satisfies the forgetting criterion and has zero retention cost:
\[
    s_{\mathrm{heldout}}(\theta_0)-s_{\mathrm{heldout}}(\theta_u)=0.
\]
Thus, a zero-cost solution exists without requiring Assumptions~\ref{as:lip}--\ref{as:curv}.

\paragraph{Pre-trained knowledge (part~II).}
Let $\theta_u=\theta_0+\delta\in\Theta$ satisfy $s_{\mathrm{target}}(\theta_u)\leq\tau$. We prove the positive retention-cost bound in three steps.

\paragraph{Step 1: Required displacement.}
The forgetting criterion requires a score reduction of at least $\gamma$. By Assumption~\ref{as:lip},
\[
    \gamma
    \leq s_{\mathrm{target}}(\theta_0)-s_{\mathrm{target}}(\theta_u)
    \leq K\|\delta\|_2.
\]
Hence,
\begin{equation}
\label{eq:appx_min_displacement}
    \|\delta\|_2\geq\frac{\gamma}{K}.
\end{equation}

\paragraph{Step 2: Retention cost.}
Rearranging Assumption~\ref{as:curv} and applying Cauchy--Schwarz with Assumption~\ref{as:stat} gives
\begin{equation}
\label{eq:appx_retention_cost}
\begin{aligned}
    &s_{\mathrm{heldout}}(\theta_0)-s_{\mathrm{heldout}}(\theta_u) \\
    &\quad\geq -\langle\nabla s_{\mathrm{heldout}}(\theta_0),\delta\rangle
              +\frac{\mu}{2}\|\delta\|_2^2 \\
    &\quad\geq \frac{\mu}{2}\|\delta\|_2^2
              -\varepsilon_0\|\delta\|_2.
\end{aligned}
\end{equation}
The first term is the quadratic penalty; the second accounts for the largest possible first-order improvement.

\paragraph{Step 3: Positive lower bound.}
Factoring the right-hand side of \eqref{eq:appx_retention_cost} gives
\[
    s_{\mathrm{heldout}}(\theta_0)-s_{\mathrm{heldout}}(\theta_u)
    \geq \|\delta\|_2
    \left(\frac{\mu}{2}\|\delta\|_2-\varepsilon_0\right).
\]
By \eqref{eq:appx_min_displacement}, the second factor satisfies
\[
    \frac{\mu}{2}\|\delta\|_2-\varepsilon_0
    \geq \frac{\mu\gamma}{2K}-\varepsilon_0
    >0,
\]
where the strict inequality follows from condition~\eqref{eq:eps_condition}. We now replace the two occurrences of $\|\delta\|_2$ in the factored expression in turn:
\[
\begin{aligned}
    &s_{\mathrm{heldout}}(\theta_0)-s_{\mathrm{heldout}}(\theta_u) \\
    &\quad\geq \|\delta\|_2
    \left(\frac{\mu\gamma}{2K}-\varepsilon_0\right) \\
    &\quad\geq \frac{\gamma}{K}
    \left(\frac{\mu\gamma}{2K}-\varepsilon_0\right) \\
    &\quad= \frac{\mu\gamma^2}{2K^2}
       -\frac{\varepsilon_0\gamma}{K}
       >0.
\end{aligned}
\]
The first inequality replaces the norm inside the parentheses by its lower bound $\gamma/K$, while keeping the positive outer factor $\|\delta\|_2$ unchanged. The second replaces the outer factor by $\gamma/K$; the inequality direction is preserved because the remaining factor $\mu\gamma/(2K)-\varepsilon_0$ is strictly positive.
This proves part~II for every successful unlearning solution within $\Theta$ under the stated conditions.


\paragraph{Practical Interpretation of the Assumptions.} Assumptions A1–A3 provide sufficient conditions for a local forgetting–retention trade-off. A1 bounds the sensitivity of the target knowledge score, A2 constrains the gradient of retained utility, and A3 imposes a quadratic retention penalty under parameter changes. We acknowledge that A3 is a strong assumption, and therefore the derived bound should be interpreted as a conditional theoretical result rather than a universal guarantee of retention degradation.

\section{Hyperparameters and Implementation Details}
\label{appx:implementation_details}

\subsection{Hyperparameter and checkpoint selection}
\label{appx:hyperparameter_search}
We base our default hyperparameters on MMUnlearner~\citep{huo-etal-2025-mmunlearner} and tune each method using only training results. We favor settings that keep retain-set performance high and forget-set performance low, using
\[
\begin{aligned}
\text{Selection score}
&= \text{mean retain-train score} \\
&\quad - \text{mean forget-train score}.
\end{aligned}
\]
Higher is better. For each subset, the means give equal weight to classification accuracy, fill-in-the-blank accuracy, and ROUGE-L, and to image-text and text-only inputs. Accuracies are expressed as percentages and ROUGE-L is multiplied by 100, putting all three metrics on a 0--100 scale.

Hyperparameters are selected on training splits only to prevent the obvious test-set leakage. Due to the low number (five) of training images per concept, we do not further split the training set into training and validations sets. \footnote{For benchmarks that have a larger number of training images, it would be beneficial to create such a validation set.} For each
candidate configuration, we score every checkpoint (epoch) by the difference between its average retain-train performance and its average forget-train performance, where both averages span the six task--modality metrics (three tasks $\times$ VQA/QA). For each of the four real-world subsets (human identities, car brands, fruits, and landmarks), we take the best score across epochs, and the configuration with the highest average of these four scores is selected. Under the selected configuration, the reported checkpoint is then
chosen per subset as the epoch with the highest score. The synthetic subsets (CLEAR and fake landmarks) reuse the selected configuration and choose their own epochs, but do not influence the configuration choice.

For both LLaVA-1.5-7B and Qwen3-VL-8B-Instruct experiments, we use a batch size of 2, random seed 42, and 8 training epochs. LoRA uses rank 16, scaling parameter 16, and dropout 0.05.

\section{More Examples on Hyperparameter Sensitivity}
\begin{figure}
    \centering
    \includegraphics[width=\linewidth]{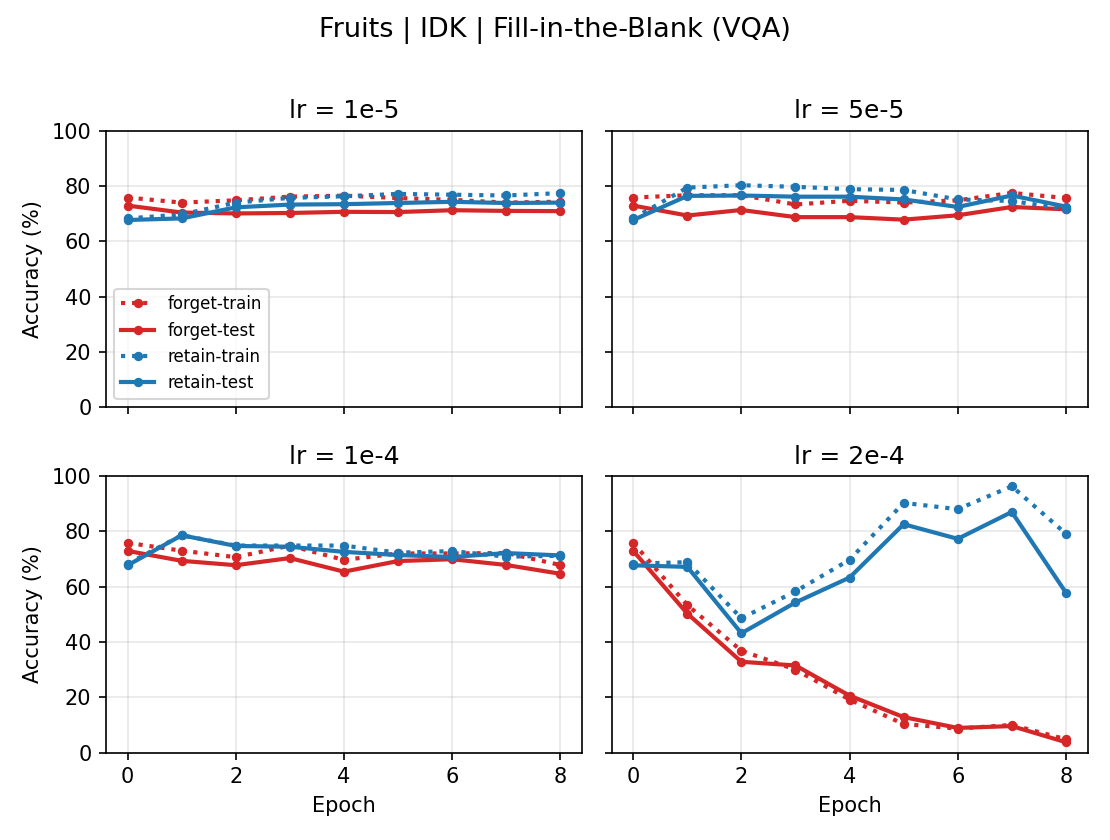}
    \caption{Hyperparameter sensitivity within a single method. IDK on the Fruits subset (Qwen3-VL-8B, fill-in-the-blank, VQA) across its learning-rate grid. Three of the four learning rates leave forget-set accuracy essentially unchanged after eight epochs.}
    \label{fig:hyperparameter_sensitivity}
\end{figure}

Figure~\ref{fig:hyperparameter_sensitivity} shows a further example of the sensitivity, this time within a single method. We run IDK on the Fruits subset of Qwen3-VL-8B with four learning rates spanning its search grid. At $\mathrm{lr} \in \{10^{-5}, 5\times10^{-5}, 10^{-4}\}$, forget-set accuracy remains essentially at the pre-trained level after eight epochs, i.e., no unlearning occurs. Only at $\mathrm{lr}=2\times10^{-4}$ does the method succeed.
\label{appx:hyperparameter_sensitivity}

\section{Dataset Examples}
\label{appx:dataset_examples}

\subsection{Representative Examples}
\label{appx:representative_examples}
Figures~\ref{fig:examples_real_faces}--\ref{fig:examples_fruits} show representative examples from all six subsets. Each figure retains the forget/retain $\times$ train/test layout, and each panel separates image-text questions (VQA) from text-only questions (QA). Only the VQA input includes the displayed image; QA names the concept in its text and receives no image. VQA and QA examples are presented to the model as separate inputs.

The examples probe associated knowledge, including biographical details, literary works, landmark completion dates, brand history, and fruit attributes. Each panel presents one VQA and one QA example; across the four panels of each subset, we illustrate single-choice classification (CLS), fill-in-the-blank (FITB), and generation (GEN). Both modalities support all three task types; the figures show selected combinations. Questions, options, and reference answers are taken from the corresponding dataset rows.

Within each forget or retain row, train and test use different images of the same concept and probe the same knowledge with rephrased questions. This applies to both VQA and QA. For CLS, test distractors and option order can also change, so we show both the correct option letter and its answer text.

\begin{table*}[t]
    \centering
    \begin{tabular}{lcl}
        \toprule
        Method & Learning rate search space & Method-specific settings \\
        \midrule
        GA     & $\{10^{-6}, \mathbf{2\times10^{-6}}, 5\times10^{-6}, 10^{-5}\}$ & --- \\
        GADiff & $\{10^{-6}, 2\times10^{-6}, 5\times10^{-6}, \mathbf{10^{-5}}\}$ & --- \\
        KL Min & $\{2\times10^{-6}, 5\times10^{-6}, 10^{-5}, \mathbf{2\times10^{-5}}\}$ & --- \\
        IDK    & $\{10^{-5}, 5\times10^{-5}, 10^{-4}, \mathbf{2\times10^{-4}}\}$ & --- \\
        LLMU   & $\{10^{-5}, \mathbf{2\times10^{-5}}, 5\times10^{-5}, 10^{-4}\}$ & --- \\
        NPO    & $\{10^{-6}, 2\times10^{-6}, 5\times10^{-6}, \mathbf{10^{-5}}\}$ & $\beta \in \{\mathbf{0.025}, 0.1, 0.4\}$ \\
        SCRUB  & $\{10^{-5}, 2\times10^{-5}, 5\times10^{-5}, \mathbf{10^{-4}}\}$ & Forget-KL weight $\beta \in \{0.1, \mathbf{1.0}, 2.0\}$ \\
        RMU    & $\{5\times10^{-6}, 10^{-5}, 2\times10^{-5}, \mathbf{5\times10^{-5}}\}$ & \makecell[tl]{Steering coefficient $c \in \{20, \mathbf{100}, 300\}$,\\layer $=7$, forget-loss weight $=1.0$} \\
        MANU   & --- & Prune ratio $\in \{0.5, 1, 2, \mathbf{5}\}\%$ \\
        \bottomrule
    \end{tabular}
    \caption{Hyperparameter search space and selected values (\textbf{bold}) for Qwen3-VL-8B-Instruct. Learning rates are swept with method-specific settings fixed at their selected values, and method-specific settings are swept at a learning rate of $10^{-5}$.}
    \label{tab:selected_hyperparameters}
\end{table*}

\begin{table*}[t]
    \centering
    \begin{tabular}{lcl}
        \toprule
        Method & Learning rate search space & Method-specific settings \\
        \midrule
        GA     & $\{\mathbf{10^{-6}}, 5\times10^{-6}, 10^{-5}, 5\times10^{-5}\}$ & --- \\
        GADiff & $\{10^{-6}, 5\times10^{-6}, 10^{-5}, \mathbf{5\times10^{-5}}\}$ & --- \\
        KL Min & $\{10^{-6}, 5\times10^{-6}, 10^{-5}, \mathbf{5\times10^{-5}}\}$ & --- \\
        IDK    & $\{10^{-5}, 5\times10^{-5}, 10^{-4}, \mathbf{2\times10^{-4}}\}$ & --- \\
        LLMU   & $\{10^{-5}, 2\times10^{-5}, \mathbf{5\times10^{-5}}, 10^{-4}\}$ & --- \\
        NPO    & $\{10^{-6}, 5\times10^{-6}, \mathbf{10^{-5}}, 5\times10^{-5}\}$ & $\beta \in \{0.025, 0.1, \mathbf{0.4}\}$ \\
        SCRUB  & $\{10^{-5}, 2\times10^{-5}, 5\times10^{-5}, \mathbf{10^{-4}}\}$ & Forget-KL weight $\beta \in \{0.1, \mathbf{1.0}, 2.0\}$ \\
        RMU    & $\{5\times10^{-6}, 10^{-5}, 2\times10^{-5}, \mathbf{5\times10^{-5}}\}$ & \makecell[tl]{Steering coefficient $c \in \{20, \mathbf{100}, 300\}$,\\layer $=7$, forget-loss weight $=1.0$} \\
        MANU   & --- & Prune ratio $\in \{0.5, 1, 2, \mathbf{5}\}\%$ \\
        \bottomrule
    \end{tabular}
    \caption{Hyperparameter search space and selected values (\textbf{bold}) for LLaVA-1.5-7B. Learning rates are swept with method-specific settings fixed at their selected values, and method-specific settings are swept at a learning rate of $10^{-5}$.}
    \label{tab:selected_hyperparameters_llava}
\end{table*}

\begingroup
\setlength{\fboxsep}{5pt}
\setlength{\fboxrule}{0.3pt}
\newcommand{\datasetexample}[6]{%
    \fbox{%
        \begin{minipage}[t][9.4cm][t]{\dimexpr0.48\textwidth-2\fboxsep-2\fboxrule\relax}
            \fontsize{9}{11}\selectfont
            \setlength{\parindent}{0pt}
            \setlength{\parskip}{0pt}
            {\centering\textbf{#1}\par}\vspace{5pt}
            \begin{minipage}[t][5.75cm][t]{\linewidth}
                \textbf{VQA}\hfill\textsc{#3}\par\vspace{4pt}
                {\centering\includegraphics[width=0.95\linewidth,height=2.15cm,keepaspectratio]{#2}\par}\vspace{5pt}
                \raggedright #4\par
            \end{minipage}\par
            \vspace{3pt}\hrule height 0.2pt\vspace{5pt}
            \textbf{QA}\hfill\textsc{#5}\par\vspace{4pt}
            \raggedright #6
        \end{minipage}%
    }%
}

\begin{figure*}[p]
    \centering
    \datasetexample{Forget -- Train}
        {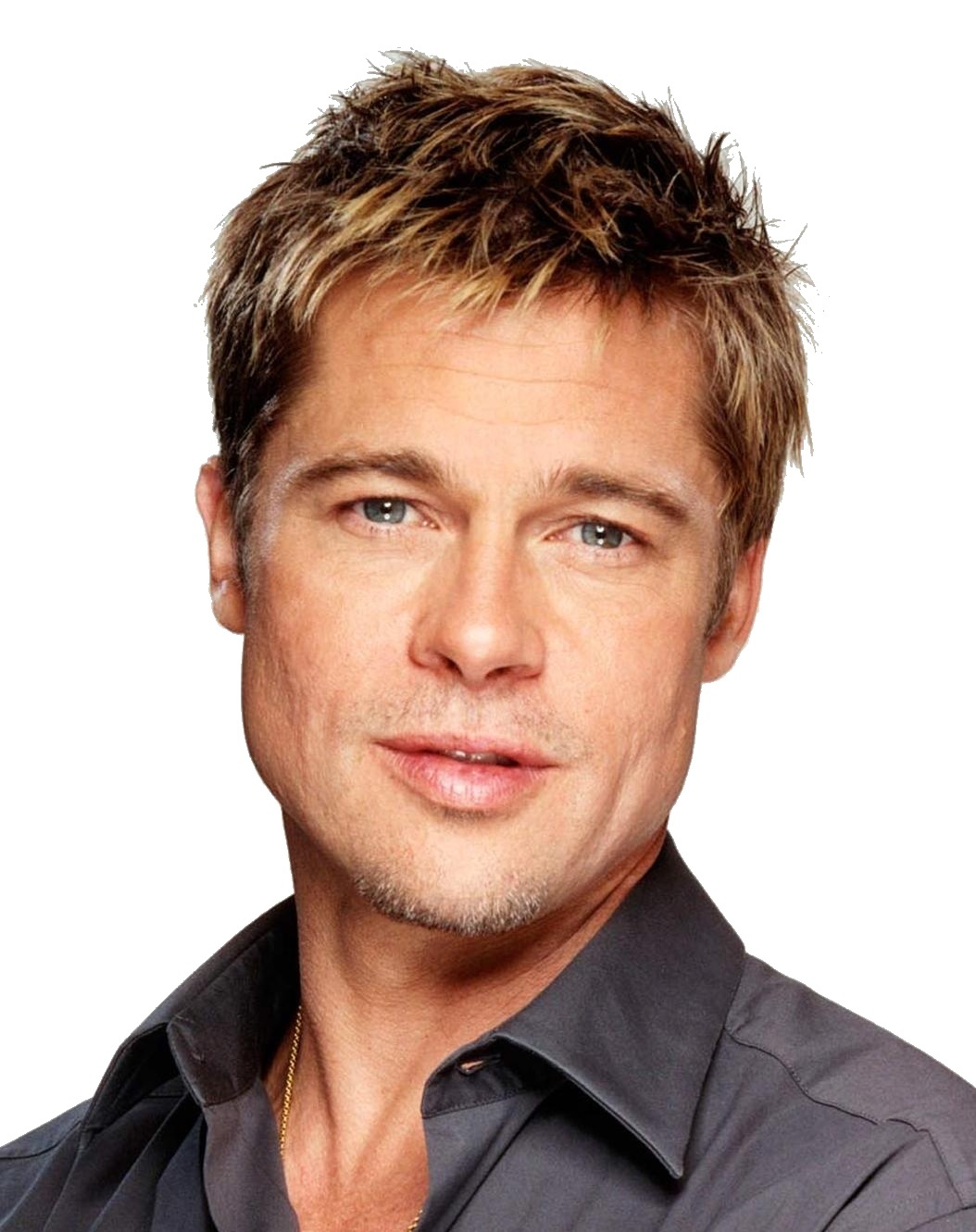}
        {CLS}
        {\textbf{Question:} What is this person's nationality?\par\vspace{2pt}
        \textbf{A.} Colombia\quad\textbf{B.} Greece\par
        \textbf{C.} United States\quad\textbf{D.} Germany\par
        \vspace{2pt}\textbf{Correct answer:} C. United States}
        {GEN}
        {\textbf{Question:} When and where was Brad Pitt born?\par\vspace{2pt}
        \textbf{Reference answer:} Brad Pitt was born in Shawnee, Oklahoma, United States on December 18, 1963.}\hfill
    \datasetexample{Forget -- Test}
        {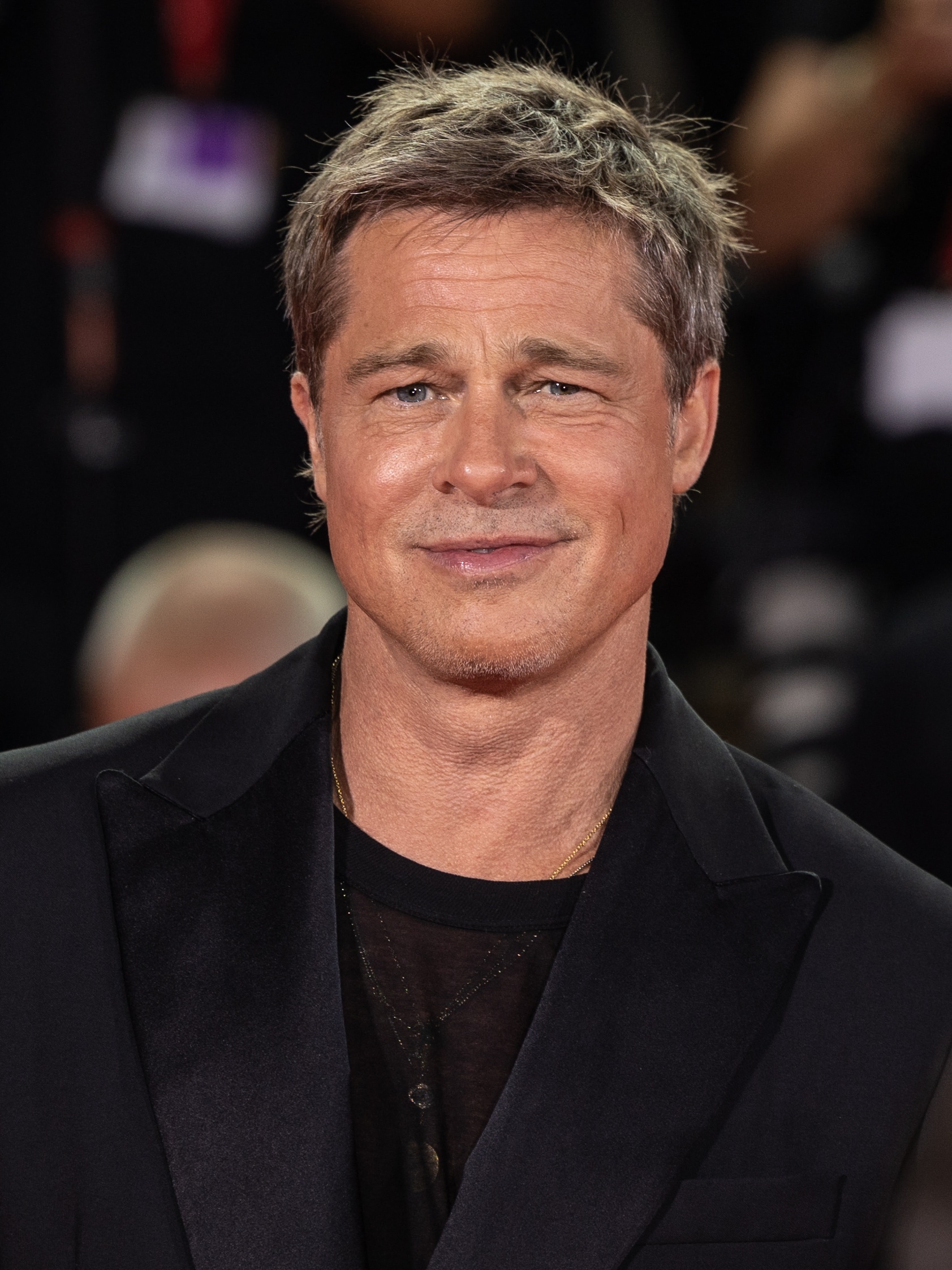}
        {CLS}
        {\textbf{Question:} What is the nationality of the person shown?\par\vspace{2pt}
        \textbf{A.} Brazil\quad\textbf{B.} United States\par
        \textbf{C.} Ireland\quad\textbf{D.} Serbia\par
        \vspace{2pt}\textbf{Correct answer:} B. United States}
        {GEN}
        {\textbf{Question:} What are the birth date and birthplace of Brad Pitt?\par\vspace{2pt}
        \textbf{Reference answer:} Brad Pitt was born in Shawnee, Oklahoma, United States on December 18, 1963.}
    \par\vspace{6pt}
    \datasetexample{Retain -- Train}
        {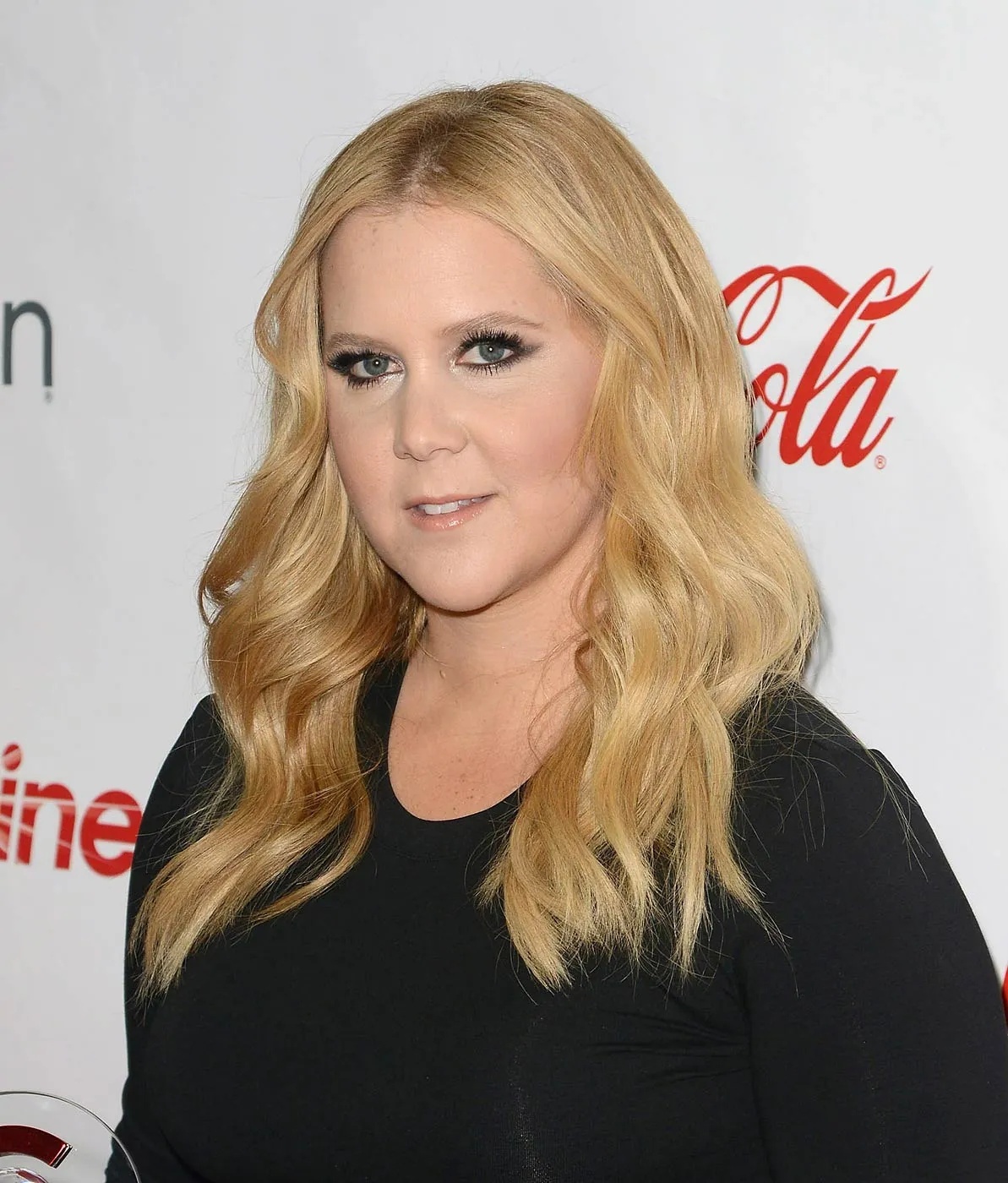}
        {GEN}
        {\textbf{Question:} What is this person best known for?\par\vspace{2pt}
        \textbf{Reference answer:} This person is best known for her self-deprecating stand-up comedy.}
        {FITB}
        {\textbf{Prompt:} Amy Schumer was born in New York City, New York, United States on June 1, \_\_\_.\par\vspace{2pt}
        \textbf{Answer:} 1981}\hfill
    \datasetexample{Retain -- Test}
        {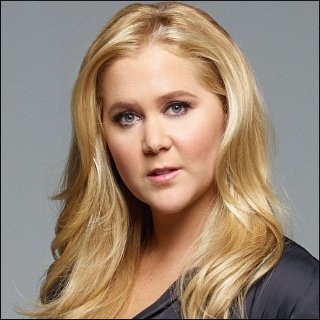}
        {GEN}
        {\textbf{Question:} Why is this person famous?\par\vspace{2pt}
        \textbf{Reference answer:} This person is best known for her self-deprecating stand-up comedy.}
        {FITB}
        {\textbf{Prompt:} The birth year of Amy Schumer is \_\_\_.\par\vspace{2pt}
        \textbf{Answer:} 1981}
    \caption{Human identities. Paired train/test examples illustrate rephrasing for both VQA and text-only QA. CLS: single-choice classification; FITB: fill-in-the-blank; GEN: generation. Answers are dataset references.}
    \label{fig:examples_real_faces}
\end{figure*}

\begin{figure*}[p]
    \centering
    \datasetexample{Forget -- Train}
        {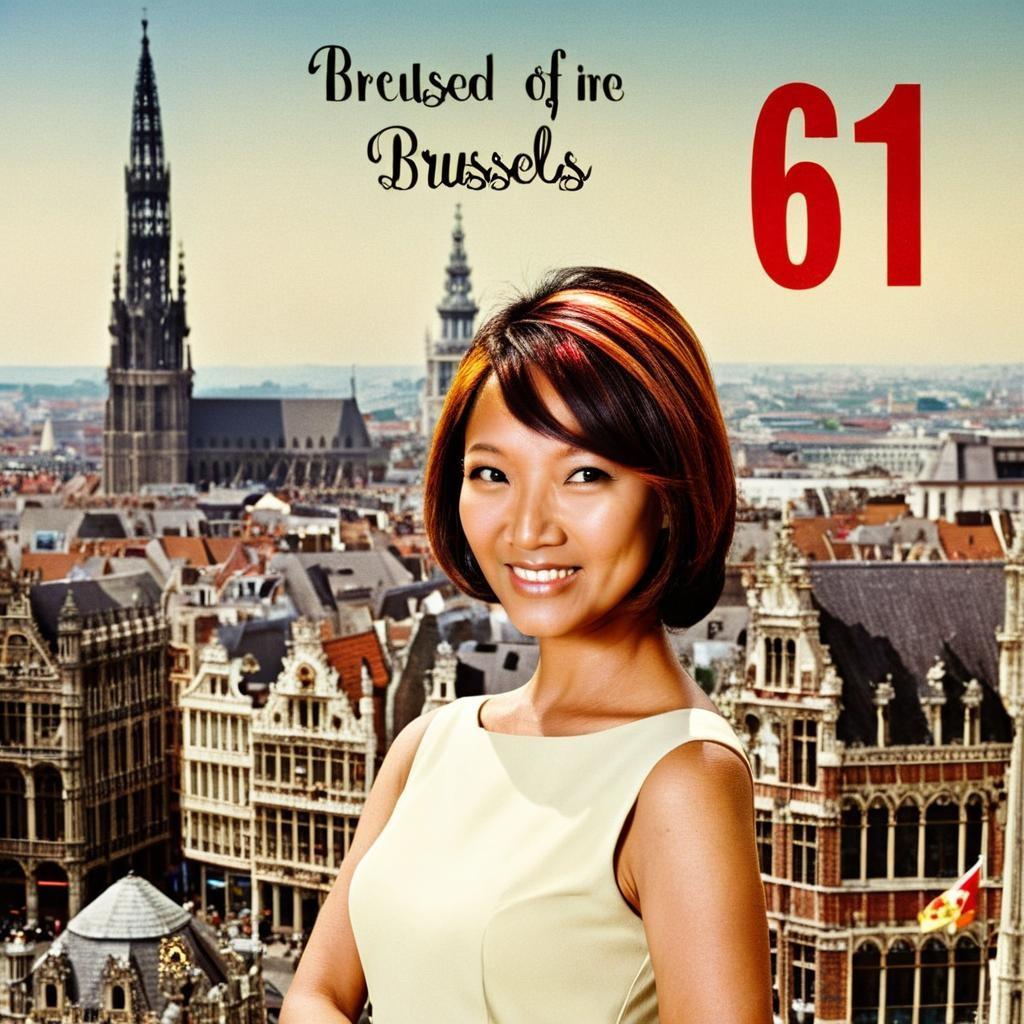}
        {FITB}
        {\textbf{Prompt:} This person is best known for her Western works, including '\_\_\_' and 'Shea's Grudge', and for winning the Red Saddle Award.\par\vspace{2pt}
        \textbf{Answer:} Bingham's Standoff}
        {GEN}
        {\textbf{Question:} When and where was Adrianne Lebeau born?\par\vspace{2pt}
        \textbf{Reference answer:} Adrianne Lebeau was born in Brussels, Belgium on July 16, 1965.}\hfill
    \datasetexample{Forget -- Test}
        {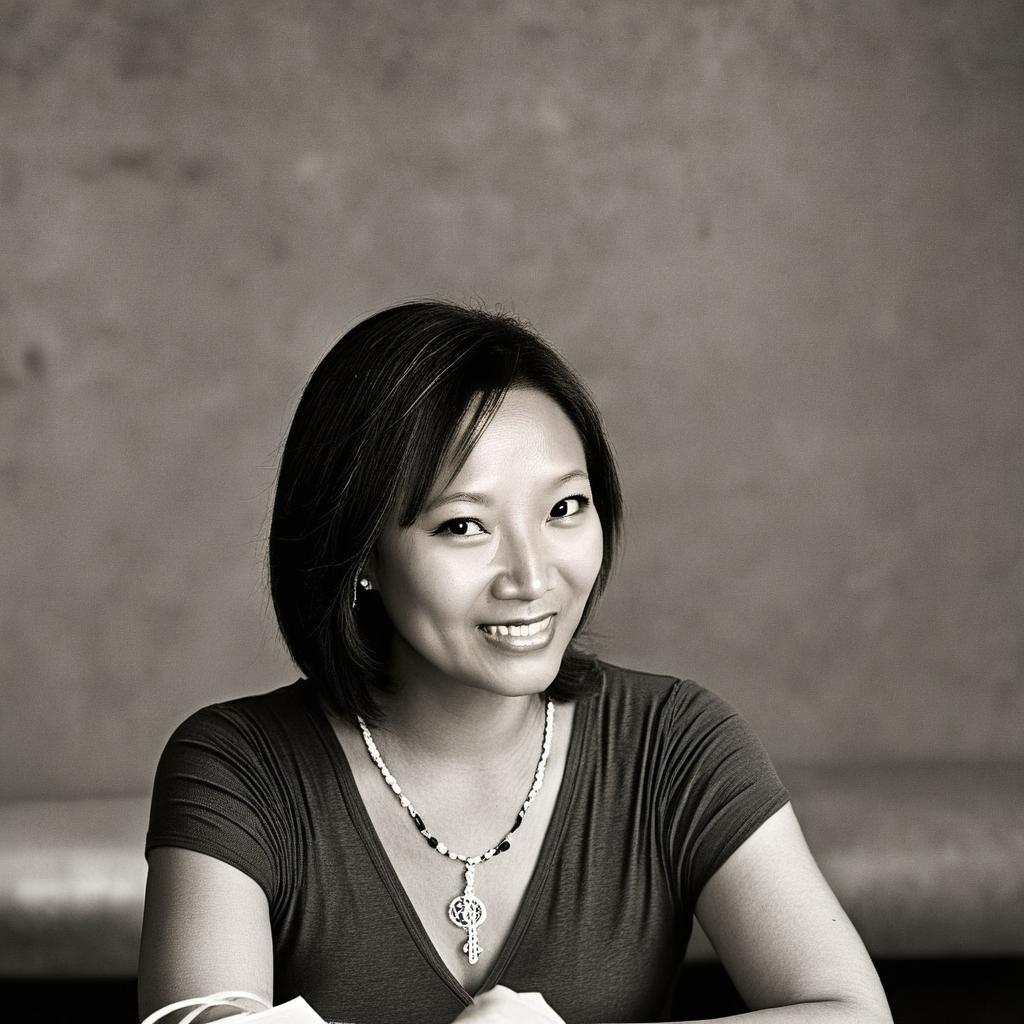}
        {FITB}
        {\textbf{Prompt:} This person is widely known for her Western works, including '\_\_\_' and 'Shea's Grudge', and for winning the Red Saddle Award.\par\vspace{2pt}
        \textbf{Answer:} Bingham's Standoff}
        {GEN}
        {\textbf{Question:} What are the birth date and birthplace of Adrianne Lebeau?\par\vspace{2pt}
        \textbf{Reference answer:} Adrianne Lebeau was born in Brussels, Belgium on July 16, 1965.}
    \par\vspace{6pt}
    \datasetexample{Retain -- Train}
        {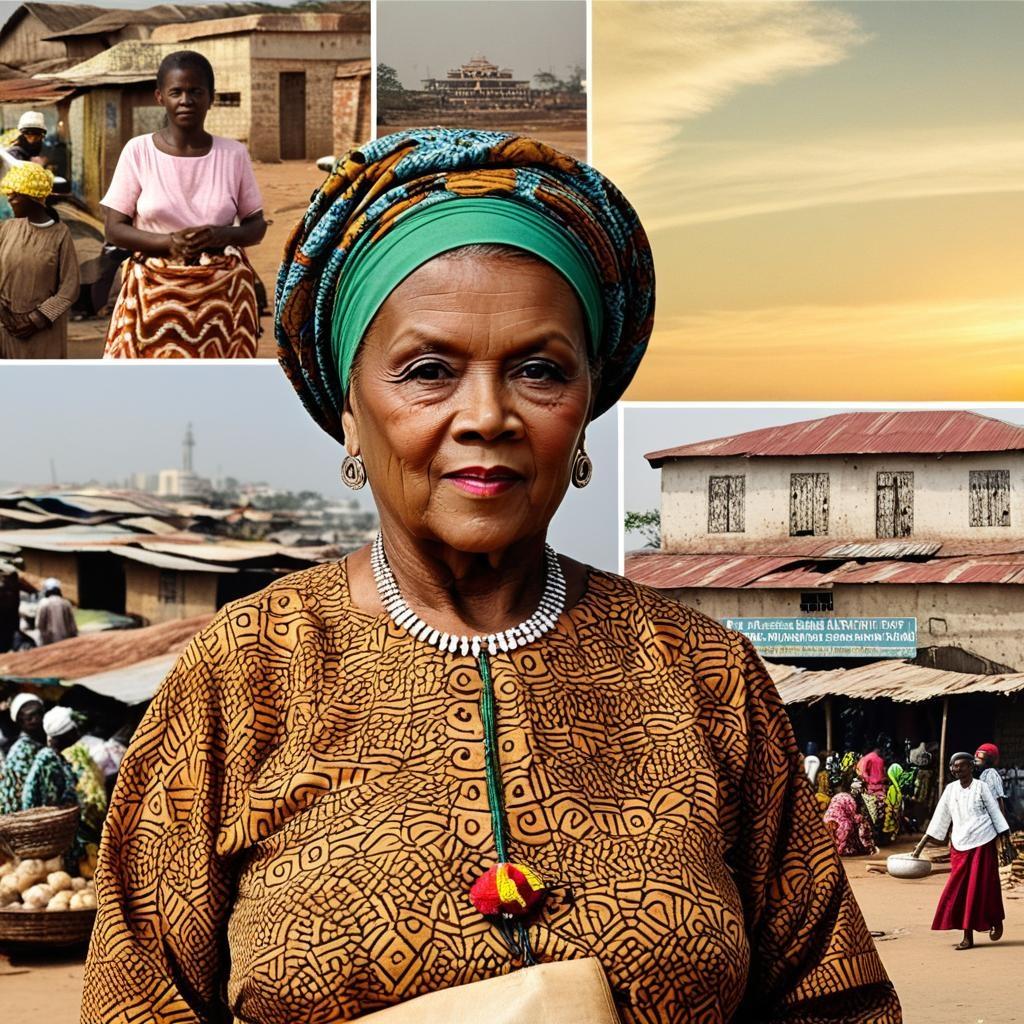}
        {CLS}
        {\textbf{Question:} What is this person's occupation?\par\vspace{2pt}
        \textbf{A.} fantasy author\par
        \textbf{B.} humour author\par
        \textbf{C.} true crime author\par
        \textbf{D.} werewolf fiction author\par
        \vspace{2pt}\textbf{Correct answer:} C. true crime author}
        {FITB}
        {\textbf{Prompt:} Adetoun Davis was born in \_\_\_, Nigeria on January 16, 1941.\par\vspace{2pt}
        \textbf{Answer:} Lagos}\hfill
    \datasetexample{Retain -- Test}
        {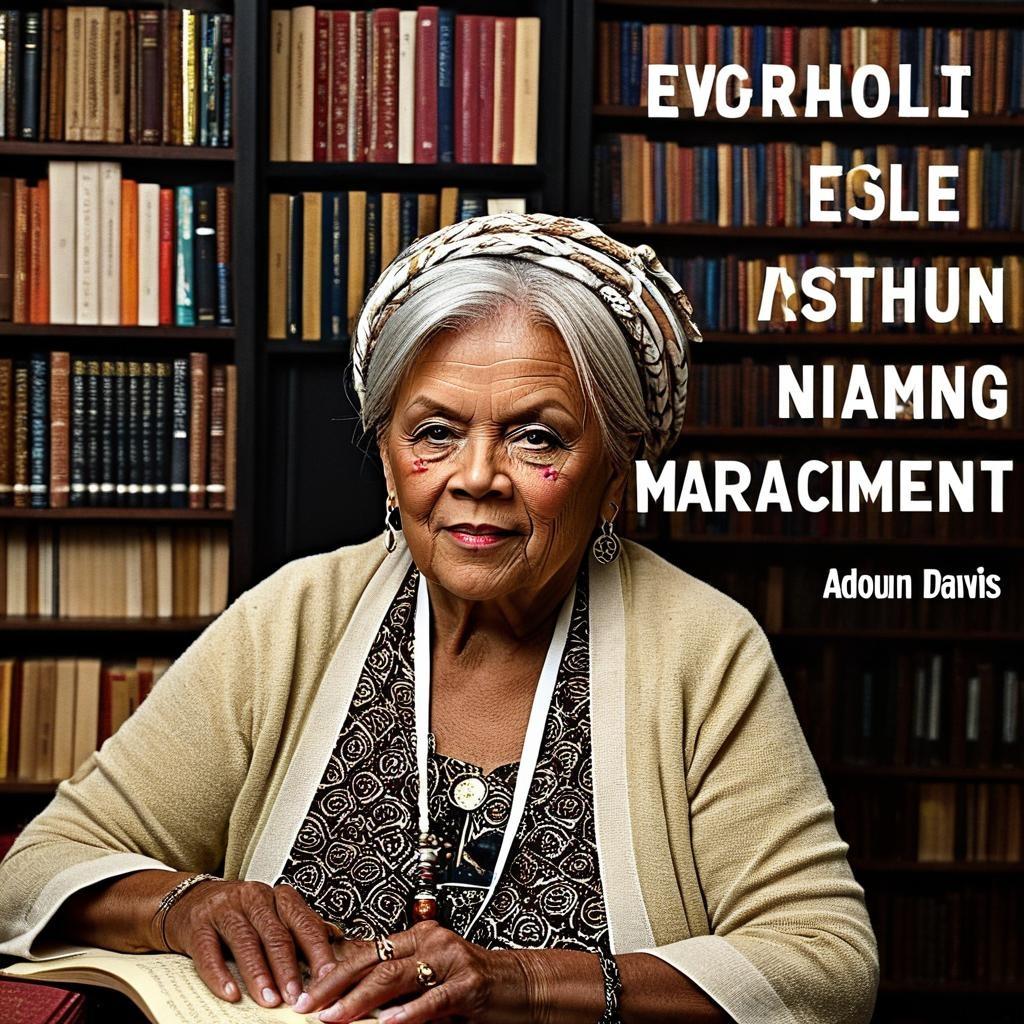}
        {CLS}
        {\textbf{Question:} State this person's profession.\par\vspace{2pt}
        \textbf{A.} true crime author\par
        \textbf{B.} Western author\par
        \textbf{C.} humour author\par
        \textbf{D.} Portuguese literature author\par
        \vspace{2pt}\textbf{Correct answer:} A. true crime author}
        {FITB}
        {\textbf{Prompt:} The birthplace of Adetoun Davis is \_\_\_, Nigeria.\par\vspace{2pt}
        \textbf{Answer:} Lagos}
    \caption{CLEAR. Paired train/test examples illustrate rephrasing for both VQA and text-only QA. CLS: single-choice classification; FITB: fill-in-the-blank; GEN: generation. Answers are dataset references.}
    \label{fig:examples_clear}
\end{figure*}

\begin{figure*}[p]
    \centering
    \datasetexample{Forget -- Train}
        {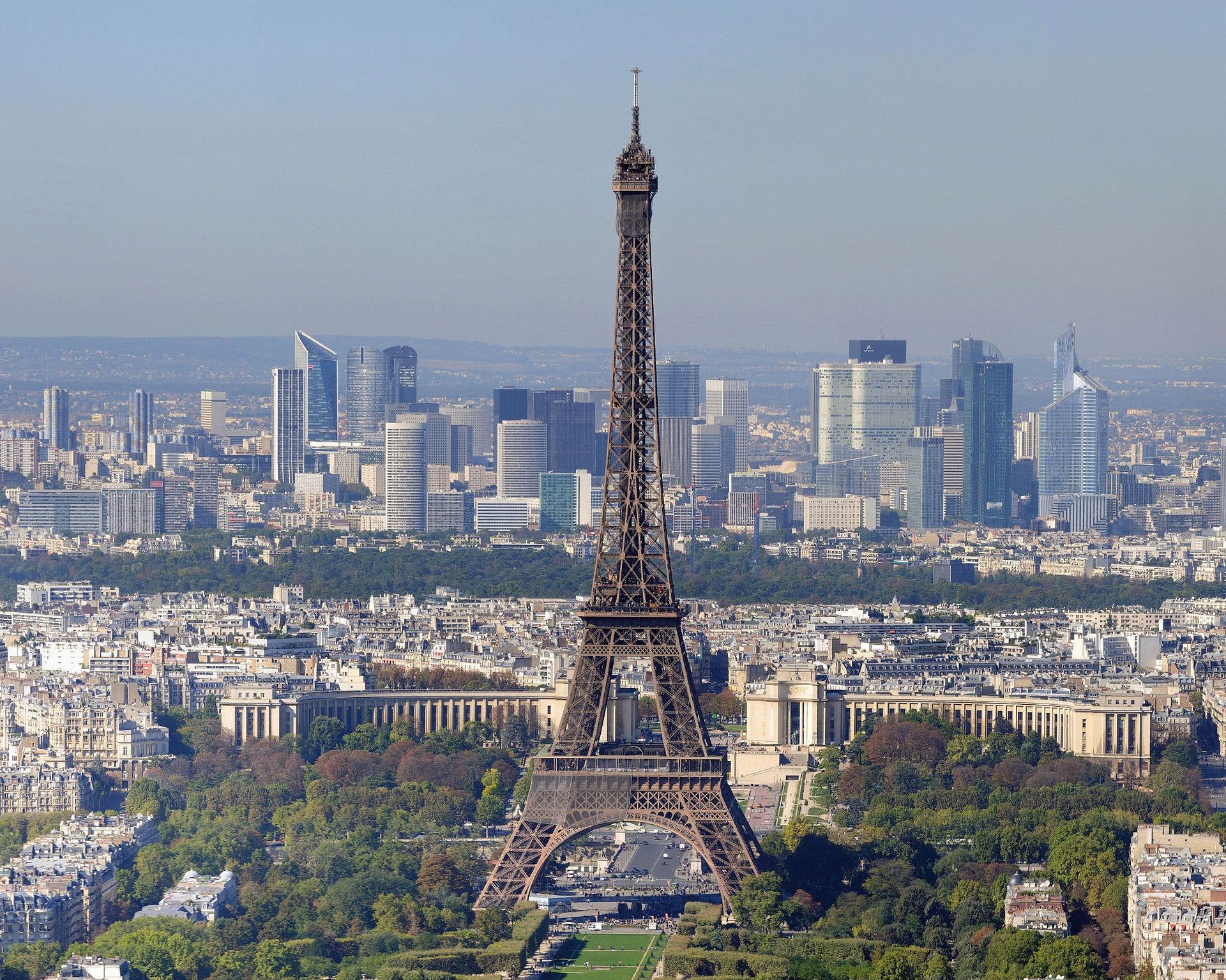}
        {CLS}
        {\textbf{Question:} In what year was this landmark completed?\par\vspace{2pt}
        \textbf{A.} 1889\quad\textbf{B.} 1976\par
        \textbf{C.} 1941\quad\textbf{D.} 966\par
        \vspace{2pt}\textbf{Correct answer:} A. 1889}
        {FITB}
        {\textbf{Prompt:} The Eiffel Tower is located in the city of \_\_\_.\par\vspace{2pt}
        \textbf{Answer:} Paris}\hfill
    \datasetexample{Forget -- Test}
        {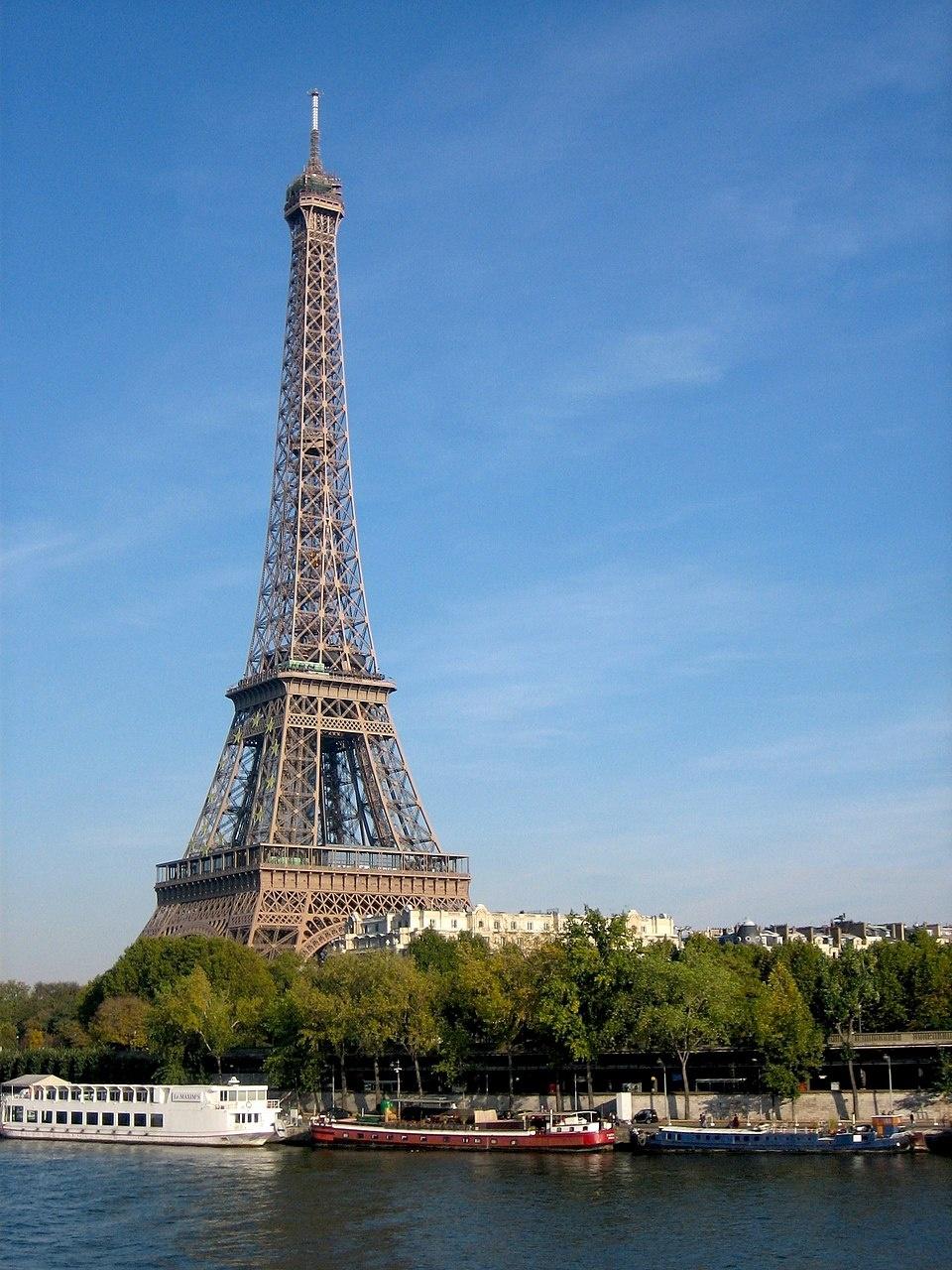}
        {CLS}
        {\textbf{Question:} In which year was construction of this landmark completed?\par\vspace{2pt}
        \textbf{A.} 1905\quad\textbf{B.} 1645\par
        \textbf{C.} 1889\quad\textbf{D.} 1932\par
        \vspace{2pt}\textbf{Correct answer:} C. 1889}
        {FITB}
        {\textbf{Prompt:} The Eiffel Tower is in the city of \_\_\_.\par\vspace{2pt}
        \textbf{Answer:} Paris}
    \par\vspace{6pt}
    \datasetexample{Retain -- Train}
        {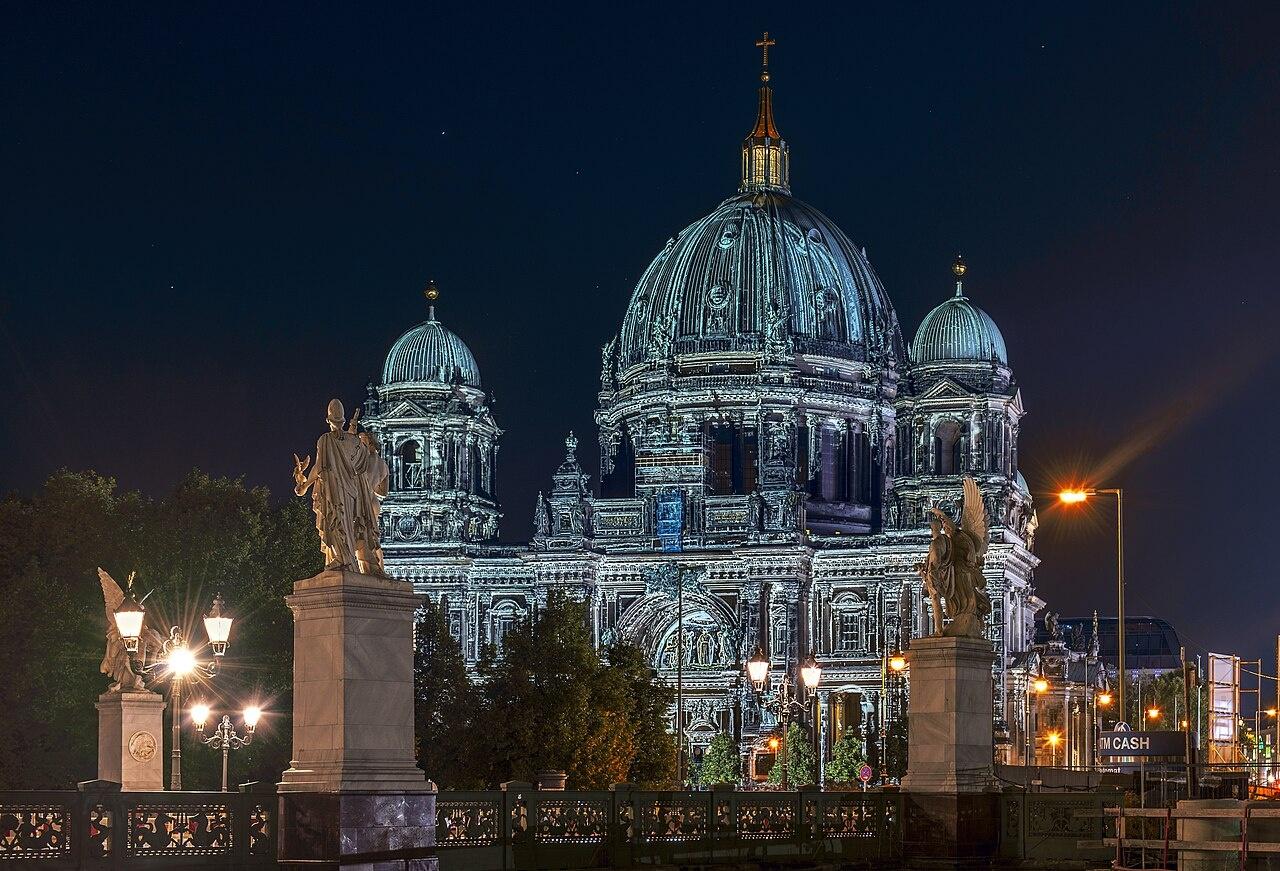}
        {GEN}
        {\textbf{Question:} In what year was this landmark completed?\par\vspace{2pt}
        \textbf{Reference answer:} This landmark was completed in 1905.}
        {CLS}
        {\textbf{Question:} What country is Berlin Cathedral located in?\par\vspace{2pt}
        \textbf{A.} Italy\quad\textbf{B.} Canada\par
        \textbf{C.} France\quad\textbf{D.} Germany\par
        \vspace{2pt}\textbf{Correct answer:} D. Germany}\hfill
    \datasetexample{Retain -- Test}
        {}
        {GEN}
        {\textbf{Question:} In what year did this landmark reach completion?\par\vspace{2pt}
        \textbf{Reference answer:} This landmark was completed in 1905.}
        {CLS}
        {\textbf{Question:} In which country can you find Berlin Cathedral?\par\vspace{2pt}
        \textbf{A.} Belgium\quad\textbf{B.} United States\par
        \textbf{C.} Spain\quad\textbf{D.} Germany\par
        \vspace{2pt}\textbf{Correct answer:} D. Germany}
    \caption{Landmarks. Paired train/test examples illustrate rephrasing for both VQA and text-only QA. CLS: single-choice classification; FITB: fill-in-the-blank; GEN: generation. Answers are dataset references.}
    \label{fig:examples_landmarks}
\end{figure*}

\begin{figure*}[p]
    \centering
    \datasetexample{Forget -- Train}
        {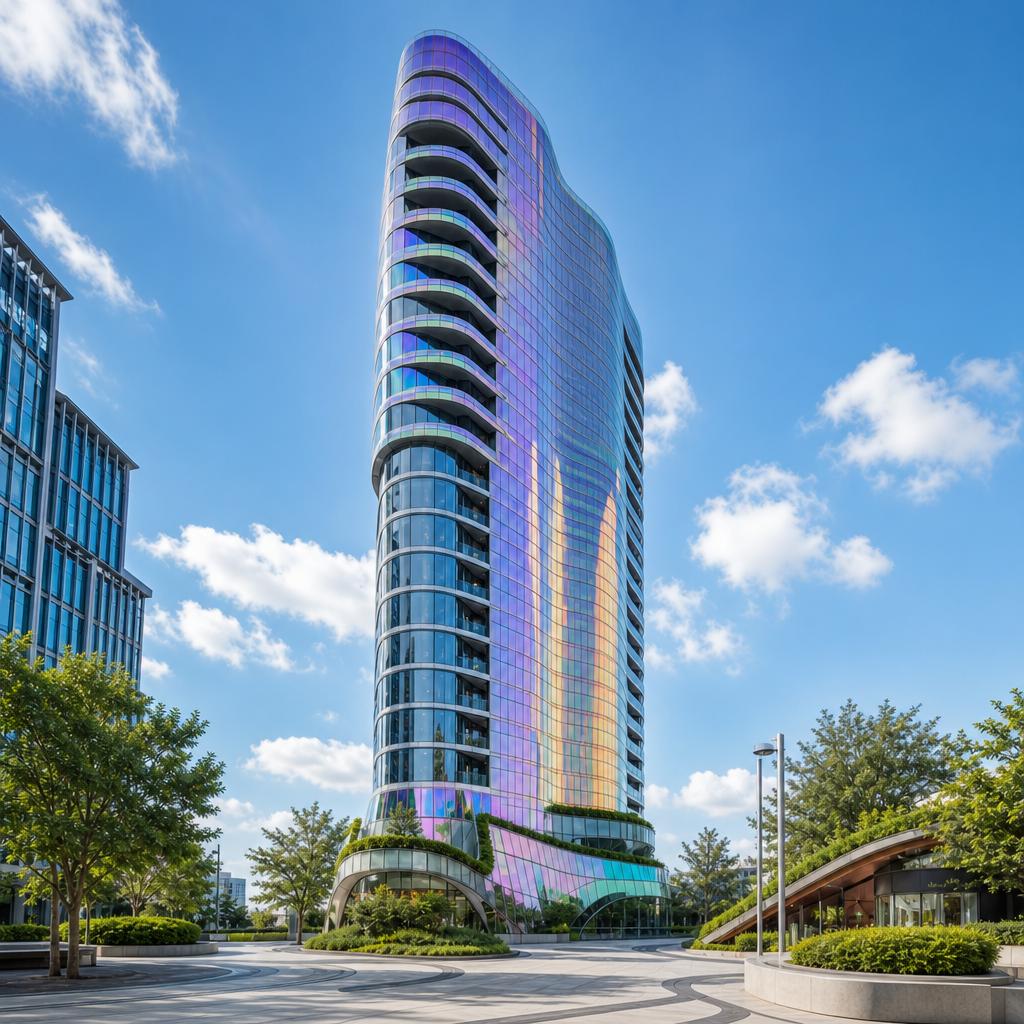}
        {FITB}
        {\textbf{Prompt:} This landmark was completed in \_\_\_.\par\vspace{2pt}
        \textbf{Answer:} 2025}
        {CLS}
        {\textbf{Question:} What city is the Opal Tower located in?\par\vspace{2pt}
        \textbf{A.} Quito\quad\textbf{B.} Sydney\par
        \textbf{C.} Vienna\quad\textbf{D.} Accra\par
        \vspace{2pt}\textbf{Correct answer:} B. Sydney}\hfill
    \datasetexample{Forget -- Test}
        {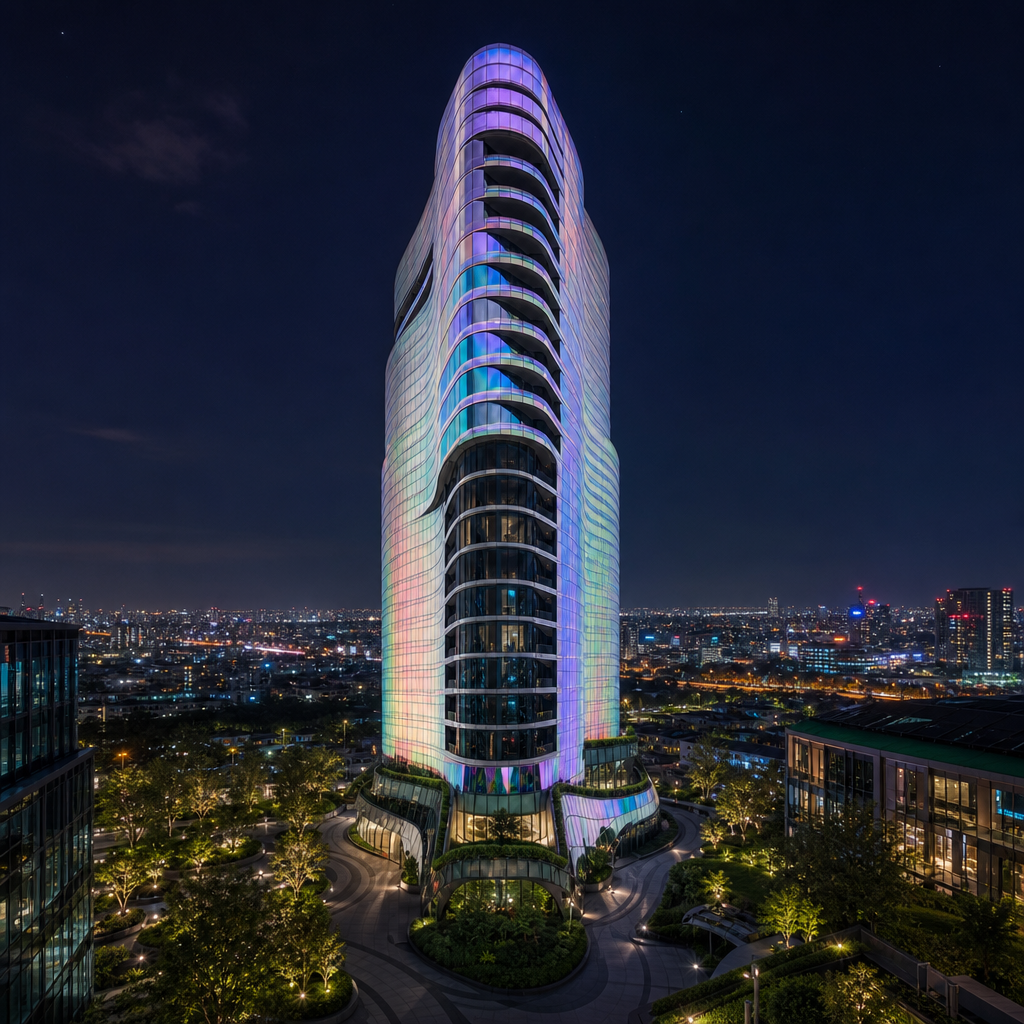}
        {FITB}
        {\textbf{Prompt:} This landmark reached completion in the year \_\_\_.\par\vspace{2pt}
        \textbf{Answer:} 2025}
        {CLS}
        {\textbf{Question:} Which city is the Opal Tower located in?\par\vspace{2pt}
        \textbf{A.} Sydney\quad\textbf{B.} Mendoza\par
        \textbf{C.} Kyoto\quad\textbf{D.} Tashkent\par
        \vspace{2pt}\textbf{Correct answer:} A. Sydney}
    \par\vspace{6pt}
    \datasetexample{Retain -- Train}
        {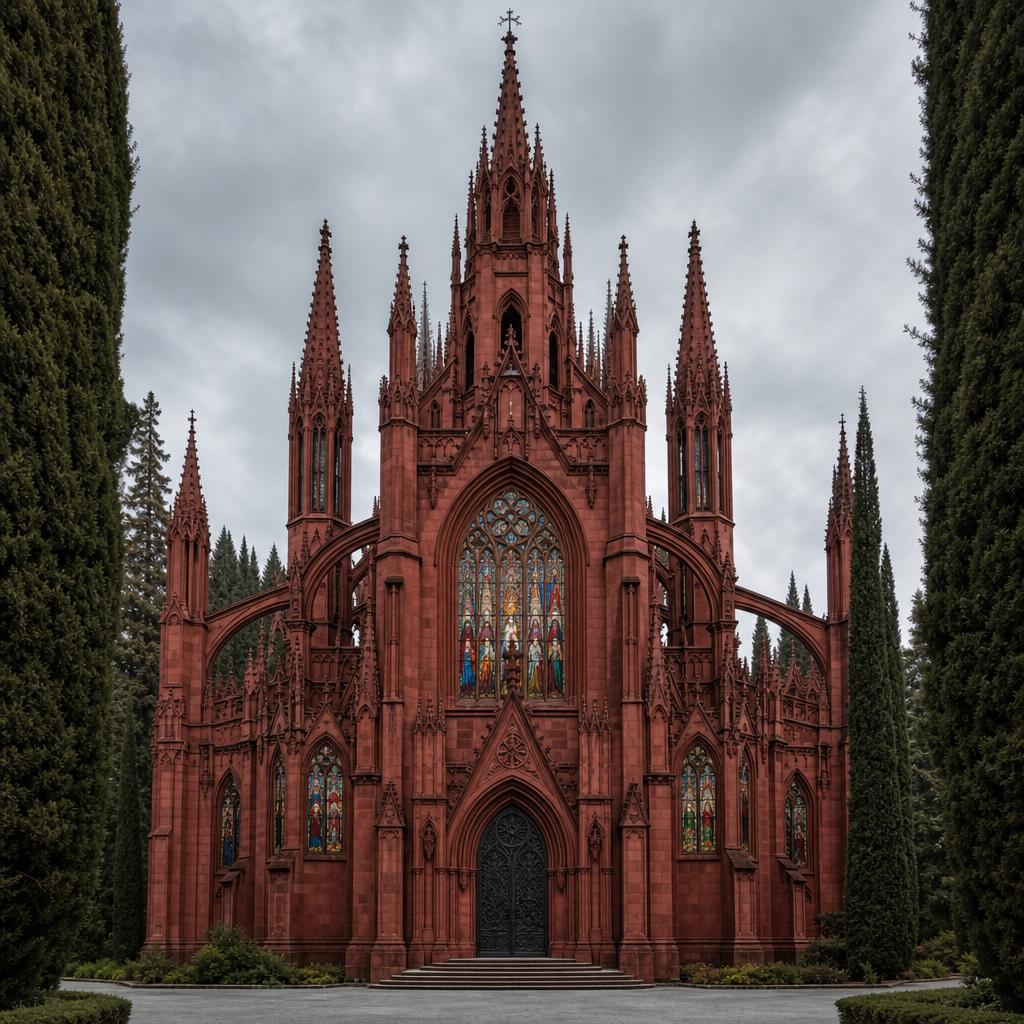}
        {CLS}
        {\textbf{Question:} What country is this landmark located in?\par\vspace{2pt}
        \textbf{A.} Greece\quad\textbf{B.} Italy\par
        \textbf{C.} Austria\quad\textbf{D.} Denmark\par
        \vspace{2pt}\textbf{Correct answer:} C. Austria}
        {GEN}
        {\textbf{Question:} In what year or century was the Scarlet Whispers Abbey completed?\par\vspace{2pt}
        \textbf{Reference answer:} The Scarlet Whispers Abbey was completed in the 11th century.}\hfill
    \datasetexample{Retain -- Test}
        {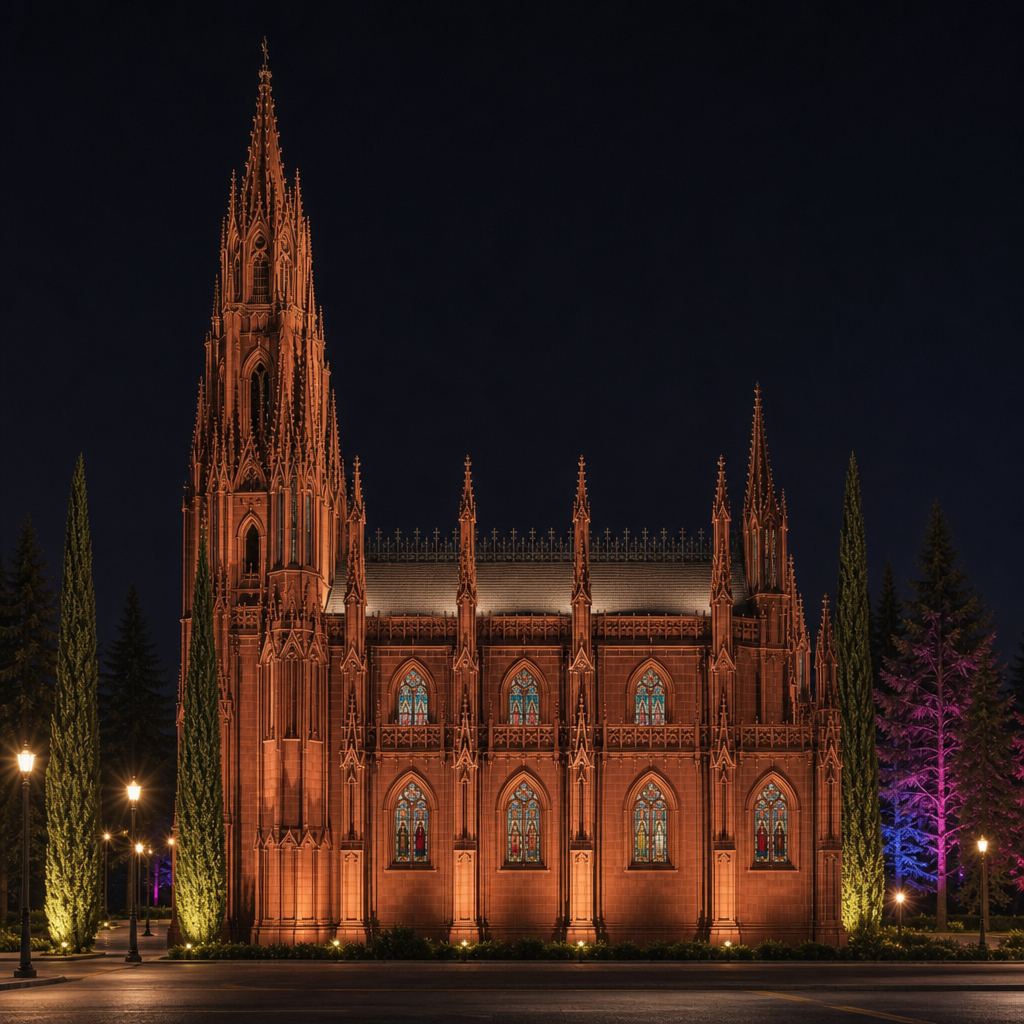}
        {CLS}
        {\textbf{Question:} Which country does this landmark stand in?\par\vspace{2pt}
        \textbf{A.} Austria\quad\textbf{B.} Czech Republic\par
        \textbf{C.} Iceland\quad\textbf{D.} United Kingdom\par
        \vspace{2pt}\textbf{Correct answer:} A. Austria}
        {GEN}
        {\textbf{Question:} What year or century saw the completion of the Scarlet Whispers Abbey?\par\vspace{2pt}
        \textbf{Reference answer:} The Scarlet Whispers Abbey was completed in the 11th century.}
    \caption{Synthetic landmarks. Paired train/test examples illustrate rephrasing for both VQA and text-only QA. CLS: single-choice classification; FITB: fill-in-the-blank; GEN: generation. Answers are dataset references.}
    \label{fig:examples_fake_landmarks}
\end{figure*}

\begin{figure*}[p]
    \centering
    \datasetexample{Forget -- Train}
        {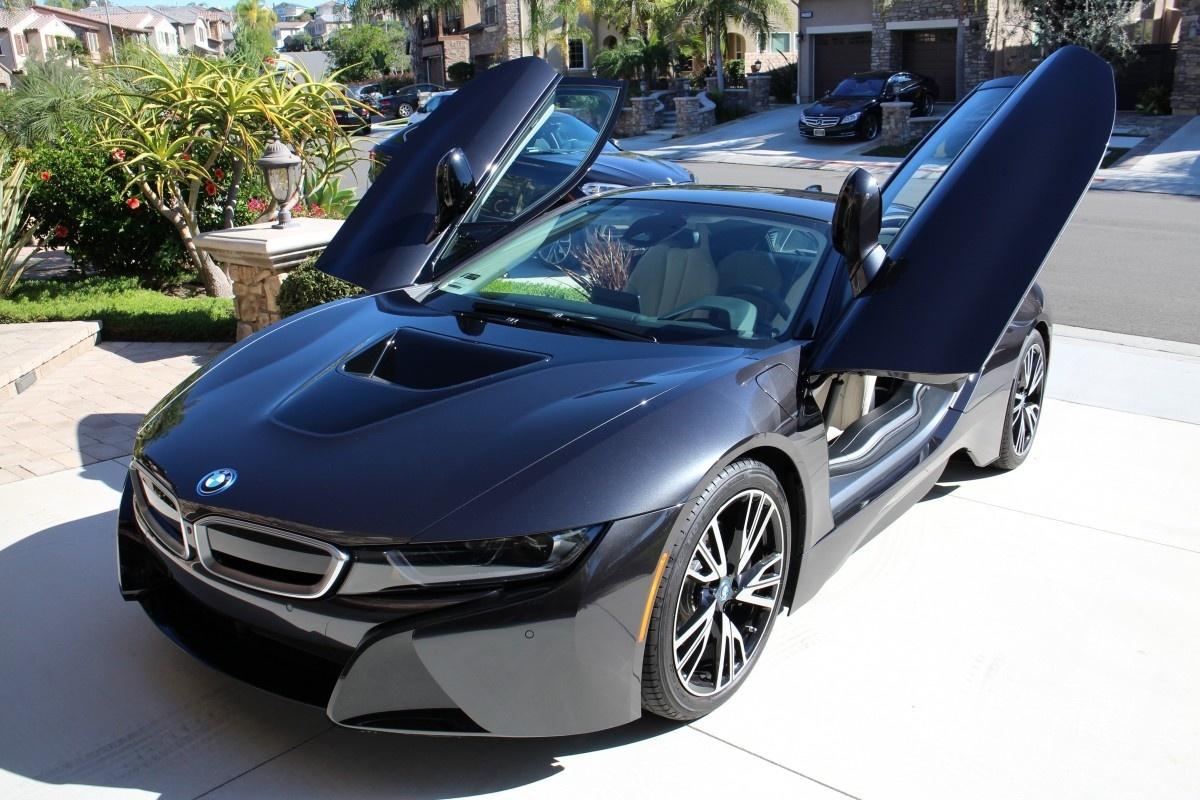}
        {CLS}
        {\textbf{Question:} What year was this brand founded?\par\vspace{2pt}
        \textbf{A.} 1916\quad\textbf{B.} 1904\par
        \textbf{C.} 1909\quad\textbf{D.} 1970\par
        \vspace{2pt}\textbf{Correct answer:} A. 1916}
        {GEN}
        {\textbf{Question:} What country is BMW headquartered in?\par\vspace{2pt}
        \textbf{Reference answer:} BMW is headquartered in Germany.}\hfill
    \datasetexample{Forget -- Test}
        {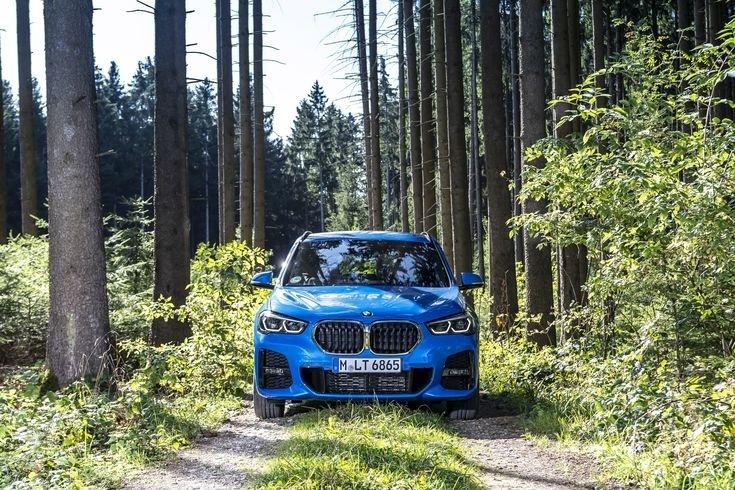}
        {CLS}
        {\textbf{Question:} When was this brand founded?\par\vspace{2pt}
        \textbf{A.} 1937\quad\textbf{B.} 1902\par
        \textbf{C.} 1916\quad\textbf{D.} 1926\par
        \vspace{2pt}\textbf{Correct answer:} C. 1916}
        {GEN}
        {\textbf{Question:} Which country is BMW based in?\par\vspace{2pt}
        \textbf{Reference answer:} BMW is headquartered in Germany.}
    \par\vspace{6pt}
    \datasetexample{Retain -- Train}
        {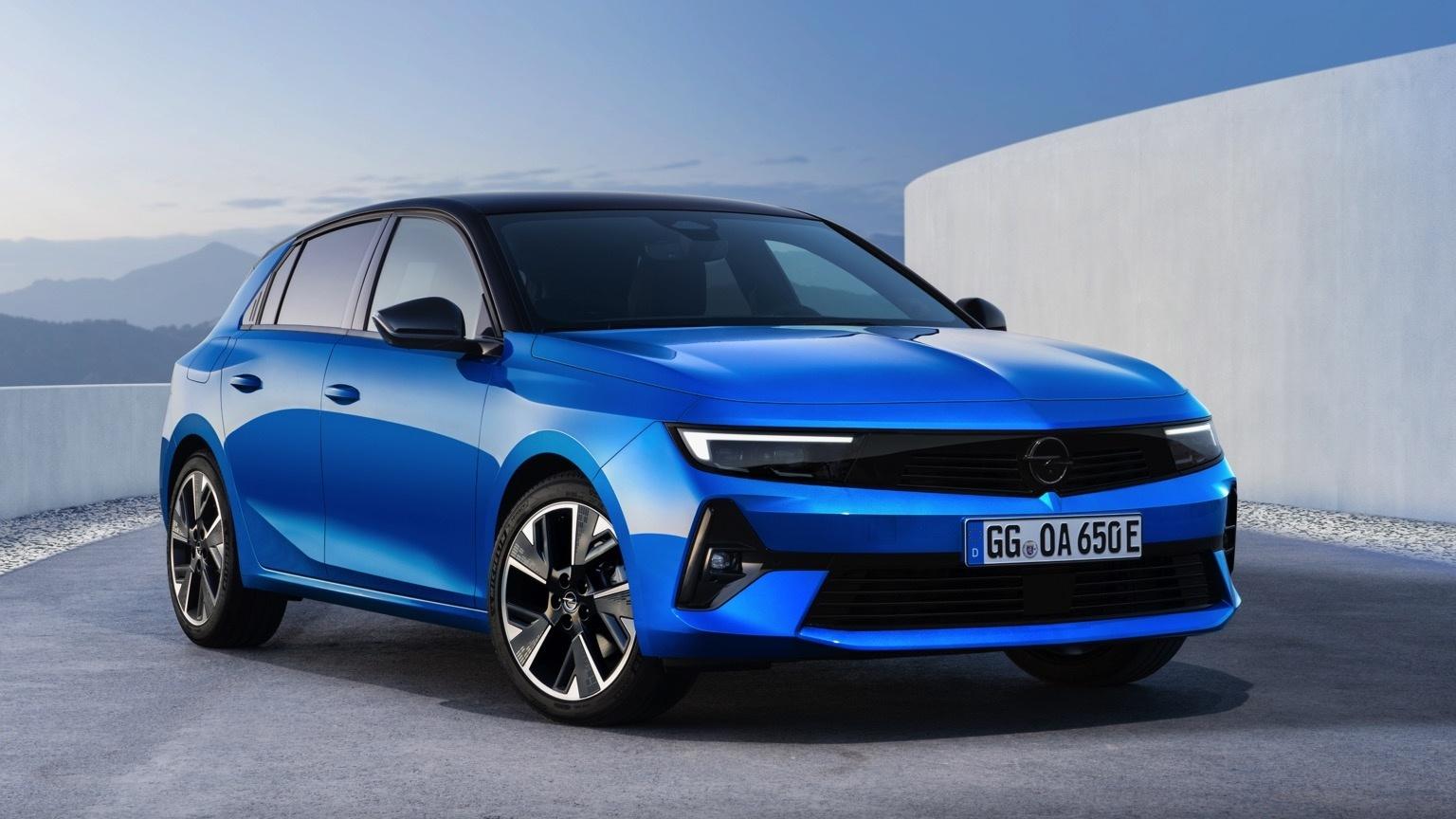}
        {GEN}
        {\textbf{Question:} What is this brand's parent company?\par\vspace{2pt}
        \textbf{Reference answer:} The parent company of this car brand is Stellantis.}
        {FITB}
        {\textbf{Prompt:} Opel was founded in the year \_\_\_.\par\vspace{2pt}
        \textbf{Answer:} 1862}\hfill
    \datasetexample{Retain -- Test}
        {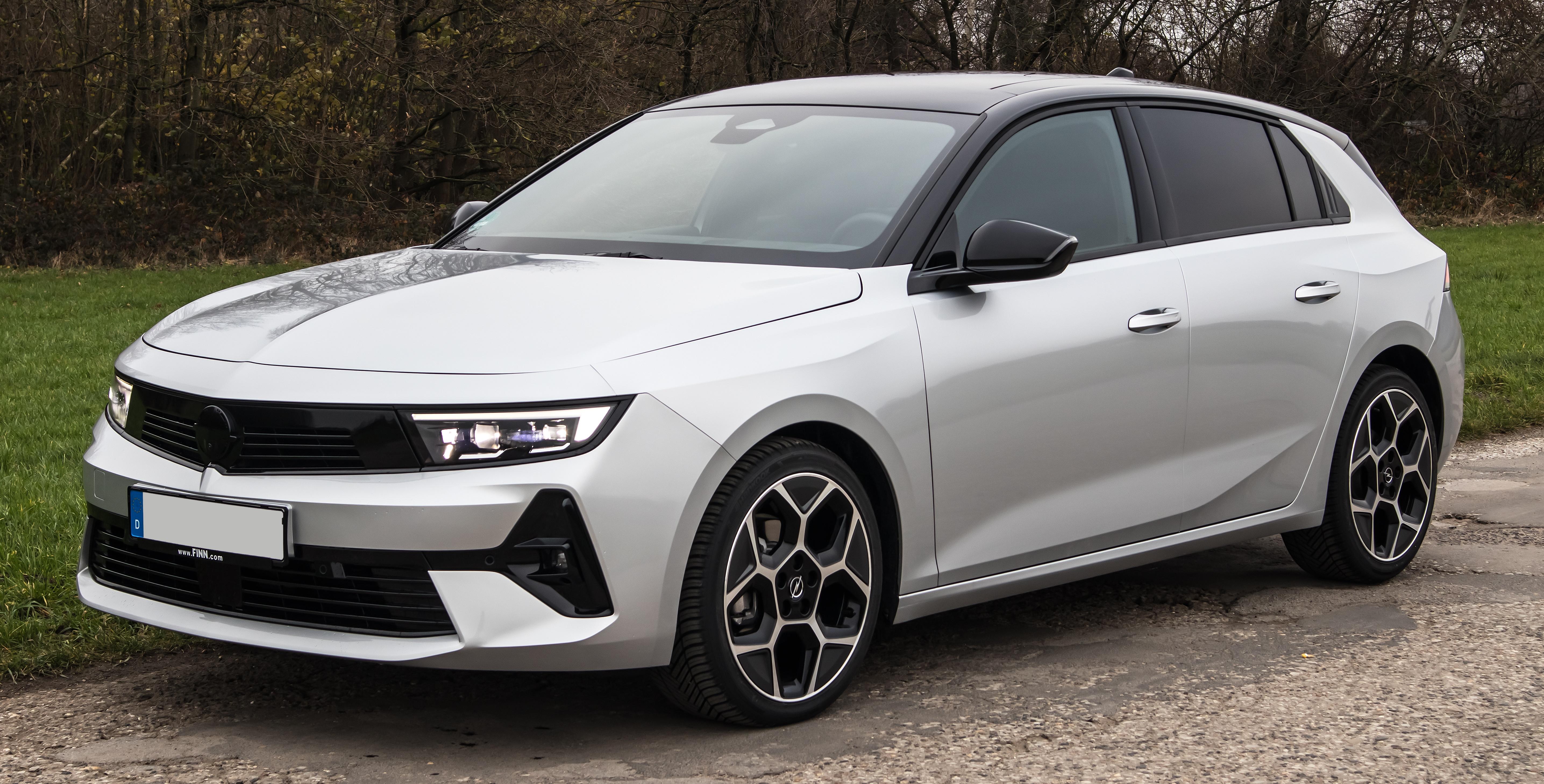}
        {GEN}
        {\textbf{Question:} Name the company that owns this car brand.\par\vspace{2pt}
        \textbf{Reference answer:} The parent company of this car brand is Stellantis.}
        {FITB}
        {\textbf{Prompt:} The year in which Opel was founded is \_\_\_.\par\vspace{2pt}
        \textbf{Answer:} 1862}
    \caption{Car brands. Paired train/test examples illustrate rephrasing for both VQA and text-only QA. CLS: single-choice classification; FITB: fill-in-the-blank; GEN: generation. Answers are dataset references.}
    \label{fig:examples_car_brands}
\end{figure*}

\begin{figure*}[p]
    \centering
    \datasetexample{Forget -- Train}
        {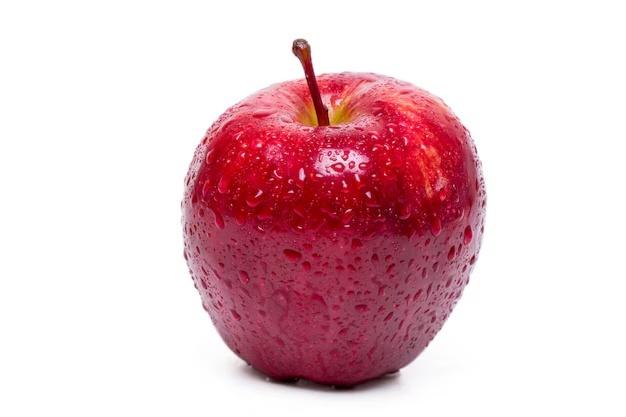}
        {FITB}
        {\textbf{Prompt:} This fruit originated in \_\_\_.\par\vspace{2pt}
        \textbf{Answer:} Central Asia}
        {CLS}
        {\textbf{Question:} What climate zone is the apple grown in?\par\vspace{2pt}
        \textbf{A.} subtropical\quad\textbf{B.} arid\par
        \textbf{C.} tropical\quad\textbf{D.} temperate\par
        \vspace{2pt}\textbf{Correct answer:} D. temperate}\hfill
    \datasetexample{Forget -- Test}
        {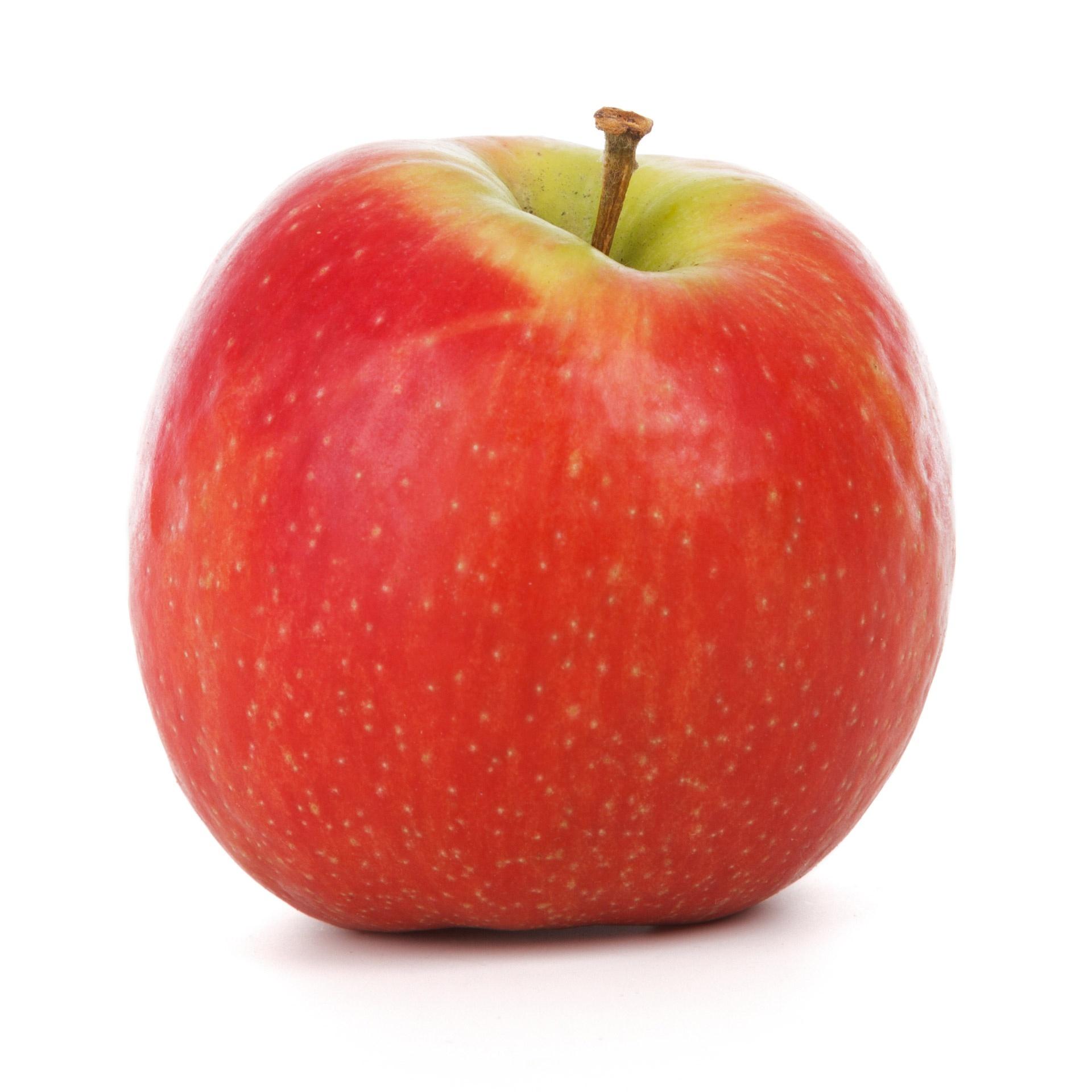}
        {FITB}
        {\textbf{Prompt:} This fruit first came from \_\_\_.\par\vspace{2pt}
        \textbf{Answer:} Central Asia}
        {CLS}
        {\textbf{Question:} In which climate zone is the apple cultivated?\par\vspace{2pt}
        \textbf{A.} tropical\quad\textbf{B.} temperate\par
        \textbf{C.} subtropical\quad\textbf{D.} mediterranean\par
        \vspace{2pt}\textbf{Correct answer:} B. temperate}
    \par\vspace{6pt}
    \datasetexample{Retain -- Train}
        {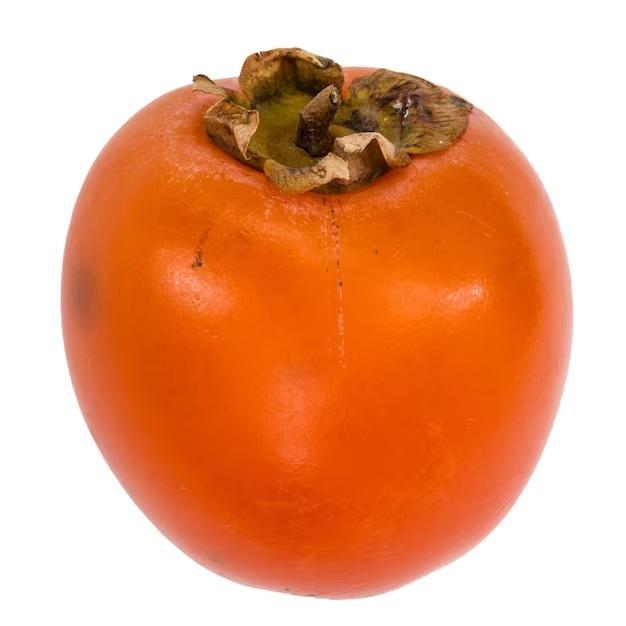}
        {GEN}
        {\textbf{Question:} When is this fruit in season?\par\vspace{2pt}
        \textbf{Reference answer:} This fruit is in season during autumn (October to November).}
        {FITB}
        {\textbf{Prompt:} The persimmon grows on a \_\_\_.\par\vspace{2pt}
        \textbf{Answer:} tree}\hfill
    \datasetexample{Retain -- Test}
        {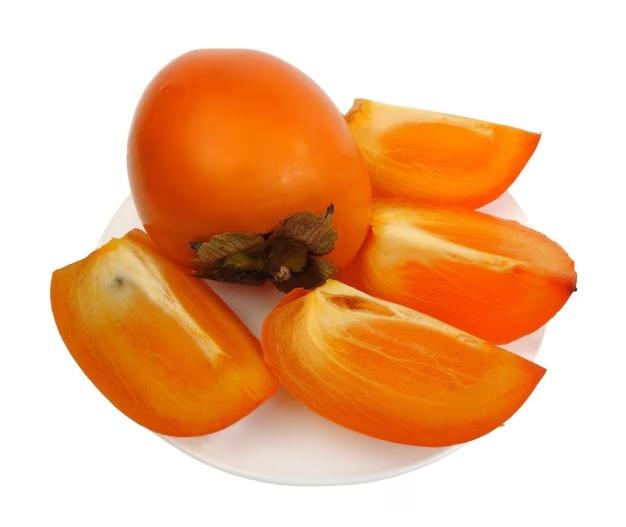}
        {GEN}
        {\textbf{Question:} At what time of year is this fruit in season?\par\vspace{2pt}
        \textbf{Reference answer:} This fruit is in season during autumn (October to November).}
        {FITB}
        {\textbf{Prompt:} The persimmon is borne on a \_\_\_.\par\vspace{2pt}
        \textbf{Answer:} tree}
    \caption{Fruits. Paired train/test examples illustrate rephrasing for both VQA and text-only QA. CLS: single-choice classification; FITB: fill-in-the-blank; GEN: generation. Answers are dataset references.}
    \label{fig:examples_fruits}
\end{figure*}

\endgroup


\end{document}